\documentclass[a4paper,11pt]{article}

\pdfoutput=1

\usepackage{natbib}
\usepackage{amsmath,amssymb,amsfonts,stmaryrd}
\usepackage[latin1]{inputenc}
\usepackage{color,rotating}
\usepackage{subscript}
\usepackage{algorithmic,algorithm2e}

\newcommand{\gcite}{\citep}
\newcommand{\gcitet}{\citet}

\newcommand{\esp}{\mathrm{E}}
\newcommand{\matr}[1]{\boldsymbol{#1}}
\newcommand{\pr}{\mathrm{p}}
\newcommand{\tr}{\mathrm{tr}}
\newcommand{\transp}{^{\sf t}}
\newcommand{\ud}{\mathrm{d}}
\newcommand{\vect}[1]{\boldsymbol{#1}}
\newcommand{\wh}{\widehat}

\graphicspath{{Figures/}}

\title{A Bayesian alternative to mutual information for the hierarchical clustering of dependent random variables}
\author{Guillaume Marrelec\textsuperscript{1,$\ast$}, Arnaud Mess{\'e}\textsuperscript{2}, Pierre Bellec\textsuperscript{3}}

\begin{document}

\maketitle

{\noindent \footnotesize \textsuperscript{1} Sorbonne Universités, UPMC Univ Paris 06, CNRS, INSERM, Laboratoire d'imagerie biomédicale (LIB), F-75013, Paris, France\\
\textsuperscript{2} Department of Computational Neuroscience, University Medical Center Hamburg-Eppendorf, Hamburg University, Hamburg, Germany\\
\textsuperscript{3} D{\'e}partement d'informatique et recherche op{\'e}rationnelle, Centre de recherche de l'institut universitaire de g{\'e}riatrie de Montr{\'e}al, Universit{\'e} de Montr{\'e}al, Montr{\'e}al, Qc, Canada\\
$\ast$ E-mail: marrelec@lib.upmc.fr}

\begin{abstract}
The use of mutual information as a similarity measure in agglomerative hierarchical clustering (AHC) raises an important issue: some correction needs to be applied for the dimensionality of variables. In this work, we formulate the decision of merging dependent multivariate normal variables in an AHC procedure as a Bayesian model comparison. We found that the Bayesian formulation naturally shrinks the empirical covariance matrix towards a matrix set a priori (e.g., the identity), provides an automated stopping rule, and corrects for dimensionality using a term that scales up the measure as a function of the dimensionality of the variables. Also, the resulting log Bayes factor is asymptotically proportional to the plug-in estimate of mutual information, with an additive correction for dimensionality in agreement with the Bayesian information criterion. We investigated the behavior of these Bayesian alternatives (in exact and asymptotic forms) to mutual information on simulated and real data. An encouraging result was first derived on simulations: the hierarchical clustering based on the log Bayes factor outperformed off-the-shelf clustering techniques as well as raw and normalized mutual information in terms of classification accuracy. On a toy example, we found that the Bayesian approaches led to results that were similar to those of mutual information clustering techniques, with the advantage of an automated thresholding. On real functional magnetic resonance imaging (fMRI) datasets measuring brain activity, it identified clusters consistent with the established outcome of standard procedures. On this application, normalized mutual information had a highly atypical behavior, in the sense that it systematically favored very large clusters. These initial experiments suggest that the proposed Bayesian alternatives to mutual information are a useful new tool for hierarchical clustering.
\par
\
\end{abstract}

\section{Introduction}  

Cluster analysis aims at uncovering natural groups of objects in a multivariate dataset [see \gcitet{Jain-2010} for a review]. In the vast variety of methods used in cluster analysis, an agglomerative hierarchical clustering (AHC) is a generic procedure that sequentially merges pairs of clusters that are most similar according to an arbitrary function called similarity measure, thereby generating a nested set of partitions, also called hierarchy \gcite{Duda-2000}. The choice of the similarity measure indirectly defines the shape of the clusters, and thus plays a critical role in the clustering process. While this choice is guided by the features of the problem at hand, it is also often restricted to a limited number of commonly used measures, such as the Euclidean distance or Pearson correlation coefficient \gcite{Dhaeseleer-2005}. In the present work, we focus on the clustering of random variables based on their mutual information, which has recently gained in popularity in cluster analysis, notably in the field of genomics \gcite{Butte-2000, Zhou_Xiaobo-2004, Dawy-2006, Priness-2007} and in functional magnetic resonance imaging (fMRI) data analysis \gcite{Stausberg-2009, Benjaminsson-2010, Kolchinsky-2014}. Mutual information is a general measure of statistical dependency derived from information theory \gcite{Shannon-1948, Kullback-1968, Cover_TM-1991}. A key feature of mutual information is its ability to capture nonlinear interactions for any type of random variables \gcite{Steuer-2002}; also of interest, it indifferently applies to univariate or multivariate variables and can thus be applied to clusters of arbitrary size. Yet, mutual information is an extensive measure that increases with variable dimensionality. In addition, $\hat{I}$, the finite-sample estimator of mutual information, suffers from a dimensionality-dependent bias (see Appendix~\ref{ann:homog}). Several authors have proposed to correct mutual information for dimensionality by using a ``normalized'' version of mutual information \gcite{Li_M-2001, Kraskov-2005, Kraskov-2009}. In the clustering literature, normalized mutual information is routinely used. However, the impact of such correction procedure has not been extensively evaluated so far.
\par
In the present paper, we consider Bayesian model-based clustering \gcite{Scott_AJ-1971, Binder-1981, Heller_KA-2005, Jain-2010} as an alternative to mutual information for the hierarchical clustering of dependent multivariate normal variables. Specifically, we derive a similarity measure by comparing two models: $\mathcal{M}_I$ where $\vect{X}_i$ and $\vect{X}_j$ are independent (i.e., the covariance between any element of $\vect{X}_i$ and any element of $\vect{X}_j$ is equal to zero), against $\mathcal{M}_D$ where the covariance between $\vect{X}_i$ and $\vect{X}_j$ can be set to any admissible
value. The proposed similarity measure is then the log Bayes factor in favor of $\mathcal{M}_D$ against $\mathcal{M}_I$ \gcite{Kass-1995}.With appropriate priors on the model parameters, we show that the similarity measure $s ( \vect{X}_i, \vect{X}_j)$ between $\vect{X}_i$ and $\vect{X}_j$ can be expressed in closed form. As will be shown below, the Bayesian formulation naturally (1) allows for clustering even when the sample covariance matrix is ill-defined; (2) provides for an automated stopping rule when the clustering reached has $s ( \vect{X}_i, \vect{X}_j ) \leq 0$ for any pair of remaining clusters; (3) corrects for dimensionality using a term that scales up the measure as a function of the dimensionality of the variables; and (4) provides for a local \emph{and} global measure of similarity, in that it can be used to decide which pair of variables to cluster at each step (local level) as well as to compare different levels of the resulting hierarchy (global level). Asymptotically (i.e., when the number of samples $N \to \infty$), the similarity measure is a linear function of mutual information, with a penalization factor that is in agreement with the Bayesian information criterion (BIC) \gcite{Schwarz-1978}. In this sense, the present paper makes an explicit connection between Bayesian model comparison for the clustering of dependent random variables and mutual information. The code corresponding to the Bayesian approach is freely available online\footnote{\texttt{https://github.com/SIMEXP/arXiv-1501.05194/releases/tag/1.0}}.
\par
We evaluated an AHC procedure based on this approach with synthetic datasets. The experiment aimed to evaluate how it behaved under both its exact and asymptotic forms compared to other approaches, including raw and normalized mutual information. We finally tested the new measures on two real datasets: a toy dataset and functional magnetic resonance imaging (fMRI) data.

\section{Analysis}

In the following, we develop a Bayesian solution to the problem of clustering detailed above. We first introduce the model together with the Bayesian framework and a general expression for the similarity measure. In subsequent subsections, we derive a closed-form expression for the marginal model likelihoods under both assumptions of dependence and independence as well as exact and asymptotic expressions for the similarity measure. We then provide a description of the hierarchical agglomerative clustering algorithm resulting from the present development. We examine how the same framework can be conveniently used to compare nested partitions, that is, different levels of a hierarchy. We also deal with the issue of setting the hyperparameters. Finally, we show how the Bayesian solution can naturally provide for an automatic stopping rule.

\subsection{Bayesian model comparison}

Let $\vect{X}$ be a $D$-dimensional multivariate normal variable with known mean\footnote{For the sake of simplicity, we assumed a known mean in the following theoretical development. If the mean is unknown (as will be the case in the simulation and real data sections), this development is still valid, with $N$ replaced by $N - 1$ and $\matr{S}$ by $$\sum_{ n = 1 } ^ N ( \vect{x_n} - \vect{m} ) ( \vect{x_n} - \vect{m} ) \transp,$$ where $\vect{m}$ is the sample mean $$\vect{m} = \frac{ 1 } { N } \sum_{ n = 1 } ^ N \vect{x}_n.$$} $\vect{\mu}$ and unknown covariance matrix $\matr{\Sigma}$. Define $\vect{X}_i$ and $\vect{X}_j$ as two disjoint subvectors of $\vect{X}$ (of size $D_i$ and $D_j$, respectively), and $\vect{X}_{ i \cup j }$ as their union (of size $D_{ i \cup j } = D_i + D_j$). Assume that we have to decide whether we should cluster $\vect{X}_i$ and $\vect{X}_j$ based on their level of dependence. To this end, consider two competing models ${\mathcal M}_I$ and ${\mathcal M}_D$. In ${\mathcal M}_I$, $\vect{X}_i$ and $\vect{X}_j$ are independent and the distribution of $\vect{X}_{ i \cup j }$ can be decomposed as the product of the marginal distributions of $\vect{X}_i$ and $\vect{X}_j$. Under such condition, $\matr{\Sigma}_{ i \cup j }$, the restriction of $\matr{\Sigma}$ to $\vect{X}_{ i \cup j }$, is block diagonal with blocks $\matr{\Sigma}_i$ and $\matr{\Sigma}_j$, the restrictions of $\matr{\Sigma}$ to $\vect{X}_i$ and $\vect{X}_j$, respectively. In ${\mathcal M}_D$, by contrast, $\vect{X}_i$ and $\vect{X}_j$ are dependent. Given a dataset $\{ \vect{x}_1, \dots, \vect{x}_N \}$ of $N$ independent and identically distributed (i.i.d.) realizations of $\vect{X}$ and $\matr{S}$ the corresponding sample sum-of-square matrix
$$\matr{S} = \sum_{ n = 1 } ^ N ( \vect{x_n} - \vect{\mu} ) ( \vect{x_n} - \vect{\mu} ) \transp,$$
we propose to quantify the similarity between $\vect{X}_i$ and $\vect{X}_j$ as the log Bayes factor, that is, the log ratio of the marginal model likelihoods of ${\mathcal M}_D$ versus ${\mathcal M}_I$ 
\begin{equation} \label{eq:s}
 s ( \vect{X}_i, \vect{X}_j ) =  \ln \frac{ \pr ( \matr{S}_{i \cup j} | {\mathcal
M}_D ) } { \pr ( \matr{S}_{i \cup j} | {\mathcal M}_I ) }.
\end{equation}
Each marginal model likelihood can be expressed as an integral over the model parameters as described below.

\subsection{Marginal model likelihood under the hypothesis of dependence}

Let us first calculate $P ( \matr{S}_{i \cup j} | {\mathcal M}_D )$, the marginal model likelihood under the hypothesis of dependence. Expressing this quantity as a function of the model parameters yields
\begin{equation} \label{eq:dep:margth}
 \pr ( \matr{S}_{i \cup j} | {\mathcal M}_D ) = \int \pr ( \matr{S}_{i \cup j} | {\mathcal M}_D, \matr{\Sigma}_{i \cup j} ) \, \pr ( \matr{\Sigma}_{i \cup j} | {\mathcal M}_D ) \, \ud \matr{\Sigma}_{i \cup j}.
\end{equation}
Calculation of the integral requires to know the likelihood $\pr ( \matr{S}_{i \cup j} | {\mathcal M}_D, \matr{\Sigma}_{i \cup j} )$ and the prior distribution $\pr ( \matr{\Sigma}_{i \cup j} | {\mathcal M}_D )$ of the covariance matrix under ${\mathcal M}_D$. With multivariate normal data, $\matr{S}_{i \cup j}$ given $\matr{\Sigma}_{i \cup j}$ is Wishart distributed with $N$ degrees of
freedom and scale matrix $\matr{\Sigma}_{i \cup j}$ \cite[Corollary~7.2.2]{Anderson_TW-2003}, leading to the following likelihood
\begin{equation} \label{eq:dep:vrais}
 \pr ( \matr{S}_{i \cup j} | {\mathcal M}_D, \matr{\Sigma}_{i \cup j} ) = \frac{ | \matr{S}_{ i \cup j } | ^ { \frac{ N - D_{i \cup j} - 1 } { 2 } } } { Z ( D_{i \cup j}, N ) } | \matr{\Sigma}_{i \cup j} | ^ { - \frac{ N } { 2 } } \exp \left[ - \frac{ 1 } { 2 } \tr ( \matr{\Sigma}_{i \cup j} ^ { - 1 } \matr{S}_{i \cup j} ) \right],
\end{equation}
where $Z ( d, n )$ is the inverse of the normalization constant
\begin{equation*}
 Z ( d, n ) = 2 ^ { \frac{ n d } { 2 } } \pi ^ { \frac{ d ( d - 1 ) } { 4 } } \prod_{ d' = 1 } ^ d \Gamma \left( \frac{ n + 1 - d' } { 2 } \right).
\end{equation*}
As to the prior distribution, this quantity is here set as a conjugate prior, namely an inverse-Wishart distribution with $\nu_{ i \cup j }$ degrees of freedom and inverse scale matrix $\matr{\Lambda}_{ i \cup j }$ \cite[\S3.6]{Gelman-2004b}
\begin{equation} \label{eq:dep:prior}
 \pr ( \matr{\Sigma}_{i \cup j} | {\mathcal M}_D ) = \frac{ | \matr{\Lambda}_{ i \cup j } | ^ { \frac{ \nu_{ i \cup j } } { 2 } } } { Z ( D_{i \cup j}, \nu_{i \cup j} ) } | \matr{\Sigma}_{i \cup j} | ^ { - \frac{ \nu_{ i \cup j } + D_{ i \cup j } + 1 } { 2 } } \exp \left[ - \frac{ 1 } { 2 } \tr ( \matr{\Sigma}_{i \cup j} ^ { - 1 } \matr{\Lambda}_{i \cup j} ) \right].
\end{equation}
Bringing Equations (\ref{eq:dep:vrais}) and (\ref{eq:dep:prior}) together into Equation~(\ref{eq:dep:margth}) yields for the marginal model likelihood
\begin{eqnarray*}
 \pr ( \matr{S}_{i \cup j} | {\mathcal M}_D ) & = & \frac{ | \matr{\Lambda}_{ i \cup j } | ^ { \frac{ \nu_{ i \cup j } } { 2 } } | \matr{S}_{ i \cup j } | ^ { \frac{ N - D_{i \cup j} - 1 } { 2 } } } { Z ( D_{i \cup j}, N ) \,  Z ( D_{i \cup j}, \nu_{i \cup j} ) } \\
 & & \quad \times \int | \matr{\Sigma}_{i \cup j} | ^ { - \frac{ N + \nu_{ i \cup j } + D_{ i \cup j } + 1 } { 2 } } \exp \left\{ - \frac{ 1 } { 2 } \tr \left[ ( \matr{\Lambda}_{i \cup j} + \matr{S}_{i \cup j} ) \matr{\Sigma}_{i \cup j} ^ { - 1 } \right] \right\} \, \ud \matr{\Sigma}_{i \cup j}.
\end{eqnarray*}
The integrand is proportional to an inverse-Wishart distribution with $N + \nu_{ i \cup j }$ degrees of freedom and scale matrix $\matr{\Lambda}_{i \cup j} + \matr{S}_{i \cup j}$, leading to
\begin{equation} \label{eq:dep:marg}
 \pr ( \matr{S}_{i \cup j} | {\mathcal M}_D ) = \frac{ | \matr{S}_{i \cup j} | ^ { \frac{ N - D_{i \cup j} - 1 } { 2 } }  }{ Z ( D_{i \cup j}, N ) } \, \frac{ Z ( D_{i \cup j}, N + \nu_{i \cup j} ) } { Z ( D_{ i \cup j }, \nu_{ i \cup j } ) } \, \frac{ | \matr{\Lambda}_{i \cup j} | ^ { \frac{ \nu_{i \cup j} } { 2 } } }{ | \matr{S}_{ i \cup j } + \matr{\Lambda}_{i \cup j} | ^ { \frac{ N + \nu_{i \cup j} } { 2 } } }.
\end{equation}

\subsection{Marginal model likelihood under the hypothesis of independence}

We can now calculate $P ( \matr{S}_{i \cup j} | {\mathcal M}_I )$, the marginal model likelihood under the hypothesis of independence. If ${\mathcal M}_I$ holds, then $\matr{\Sigma}_{ i \cup j }$ is block-diagonal with two blocks $\matr{\Sigma}_i$ and $\matr{\Sigma}_j$ the submatrix restrictions of $\matr{\Sigma}_{ i \cup j }$ to $\vect{X}_i$ and $\vect{X}_j$, respectively. Introduction of the model parameters therefore yields for the marginal likelihood
\begin{equation} \label{eq:indep:margth}
 \pr ( \matr{S}_{ i \cup j } | {\mathcal M}_I ) = \int \pr ( \matr{S}_{ i \cup j } | {\mathcal M}_I, \matr{\Sigma}_i, \matr{\Sigma}_j ) \, \pr ( \matr{\Sigma}_i, \matr{\Sigma}_j | {\mathcal M}_I ) \, \ud \matr{\Sigma}_i \, \ud \matr{\Sigma}_j.
\end{equation}
To calculate this integral, we again need to know the likelihood $\pr ( \matr{S}_{i \cup j} | {\mathcal M}_I, \matr{\Sigma}_i, \matr{\Sigma}_j )$ and the prior distribution $\pr ( \matr{\Sigma}_i, \matr{\Sigma}_j | {\mathcal M}_D )$ of the two blocks of the covariance matrix under ${\mathcal M}_I$. The likelihood is the same as for ${\mathcal M}_D$ and has the form of Equation~(\ref{eq:dep:vrais}). Furthermore, since $\matr{\Sigma}_{ i \cup j }$ is here block diagonal, we have $| \matr{\Sigma}_{i \cup j} | = | \matr{\Sigma}_i | \, | \matr{\Sigma}_j |$ and $\tr ( \matr{\Sigma}_{i \cup j} ^ { - 1 } \matr{S}_{i \cup j} ) = \tr ( \matr{\Sigma}_i ^ { - 1 } \matr{S}_i ) + \tr ( \matr{\Sigma}_j ^ { - 1 } \matr{S}_j )$, where $\matr{S}_i$ and $\matr{S}_j$ are the restrictions of $\matr{S}$ to $\vect{X}_i$ and $\vect{X}_j$, respectively. Consequently, the likelihood can be further expanded
as
\begin{equation} \label{eq:indep:vrais}
 \pr ( \matr{S}_{ i \cup j } | {\mathcal M}_I, \matr{\Sigma}_i, \matr{\Sigma}_j ) = \frac{ | \matr{S}_{ i \cup j } | ^ { \frac{ N - D_{ i \cup j } - 1 } { 2 } } } { Z ( D_{ i \cup j }, N ) }  \prod_{ k \in \{ i, j \} } | \matr{\Sigma}_k | ^ { - \frac{ N } { 2 } } \exp \left[ - \frac{ 1 } { 2 } \tr ( \matr{\Sigma}_k ^ { - 1 } \matr{S}_k ) \right].
\end{equation}
As to the prior distribution, assuming no prior dependence between $\matr{\Sigma}_i$ and $\matr{\Sigma}_j$ yields 
\begin{equation} \label{eq:indep:prior:1}
 \pr ( \matr{\Sigma}_i, \matr{\Sigma}_j | {\mathcal M}_I ) = \pr ( \matr{\Sigma}_i | {\mathcal M}_I ) \, \pr ( \matr{\Sigma}_j | {\mathcal M}_I ). 
\end{equation}
For the sake of consistency, $\pr ( \matr{\Sigma}_i | {\mathcal M}_I )$ and $\pr ( \matr{\Sigma}_j | {\mathcal M}_I )$ are set equal to $\pr ( \matr{\Sigma}_i | {\mathcal M}_D )$ and $\pr ( \matr{\Sigma}_j | {\mathcal M}_D )$, respectively, which are in turn obtained by marginalization of $\pr ( \matr{\Sigma}_{ i \cup j } | {\mathcal M}_D )$ as given by Equation~(\ref{eq:dep:prior}). For $k \in \{ i, j \}$, $\pr ( \matr{\Sigma}_k | {\mathcal M}_I )$ can be found to have an inverse-Wishart distribution with $\nu_k = \nu_{ i \cup j } - D_{ i \cup j } + D_k$ degrees of freedom and inverse scale matrix $\matr{\Lambda}_k$ the restriction of $\matr{\Lambda}_{ i \cup j }$ to $\vect{X}_k$ \cite[\S5.2]{Press_SJ-2005}
\begin{equation} \label{eq:indep:prior:2}
 \pr ( \matr{\Sigma}_k | {\mathcal M}_I ) = \frac{ | \matr{\Lambda}_k | ^ { \frac{ \nu_k } { 2 } } } { Z ( D_k, \nu_k ) } | \matr{\Sigma}_k | ^ { - \frac{ \nu_k + D_k + 1 } { 2 } } \exp \left[ - \frac{ 1 } { 2 } \tr ( \matr{\Lambda}_k \matr{\Sigma}_k ^ { - 1 } ) \right].
\end{equation}
Incorporating Equations (\ref{eq:indep:vrais}), (\ref{eq:indep:prior:1}), and (\ref{eq:indep:prior:2}) into Equation~(\ref{eq:indep:margth}) yields
\begin{eqnarray*}
 \pr ( \matr{S}_{i \cup j} | {\mathcal M}_I ) & = & \frac{ | \matr{S}_{ i \cup j } | ^ { \frac{ N - D_{ i \cup j } - 1 } { 2 } } } { Z ( D_{ i \cup j }, N ) } \prod_{ k  \in \{ i, j \} } \frac{ | \matr{\Lambda}_k | ^ { \frac{ \nu_k } { 2 } } } { Z ( D_k, \nu_k ) } \\
 & & \quad \times \int | \matr{\Sigma}_k | ^ { - \frac{ N + \nu_k + D_k + 1 } { 2 } } \exp \left\{ - \frac{ 1 } { 2 } \tr \left[ ( \matr{S}_k + \matr{\Lambda}_k ) \matr{\Sigma}_k ^ { - 1 } \right] \right\}. 
\end{eqnarray*}
Each integrand is proportional to an inverse-Wishart distribution with $N + \nu_k$ degrees of freedom and scale matrix $\matr{S}_k + \matr{\Lambda}_k$, leading to
\begin{equation} \label{eq:indep:marg}
 \pr ( \matr{S}_{i \cup j} | {\mathcal M}_I ) = \frac{ | \matr{S}_{ i \cup j } | ^ { \frac{ N - D_{ i \cup j } - 1 } { 2 } } } { Z ( D_{ i \cup j }, N ) } \prod_{ k  \in \{ i, j \} } \frac{ Z ( D_k, N + \nu_k ) } { Z ( D_k, \nu_k ) } \, \frac{ | \matr{\Lambda}_k | ^ { \frac{ \nu_k } { 2 } } } { | \matr{S}_k + \matr{\Lambda}_k | ^ { \frac{ N + \nu_k } { 2 } } }.
\end{equation}

\subsection{Log Bayes factor of dependence versus independence}

Let us now express the Bayesian similarity measure by incorporating Equations (\ref{eq:dep:marg}) and (\ref{eq:indep:marg}) into Equation~(\ref{eq:s}), yielding
\begin{equation} \label{eq:bayes:def:2a}
  s ( \vect{X}_i, \vect{X}_j ) = \left[ \sum_{ k \in \{ i, j \} } \frac{ N + \nu_k } { 2 } \ln | \matr{\Lambda}_k + \matr{S}_k  | \right] - \frac{ N + \nu_{i \cup j} } { 2 } \ln | \matr{\Lambda}_{i \cup j} + \matr{S}_{i \cup j} | + \mathrm{cst}
\end{equation}
with
\begin{eqnarray} \label{eq:bayes:def:2b}
 \mathrm{cst} & = & \frac{ \nu_{i \cup j} } { 2 } \ln | \matr{\Lambda}_{i \cup j} | + \sum_{ d = 1 } ^ { D_{ i \cup j } } \left[ \ln \Gamma \left( \frac{ N + \nu_{ i \cup j } + 1 - d } { 2 } \right) - \ln \Gamma \left( \frac{ \nu_{ i \cup j } + 1 - d } { 2 } \right) \right] \\
 & & - \sum_{ k \in \{ i, j \} } \left\{ \frac{ \nu_k } { 2 } \ln | \matr{\Lambda}_k | + \sum_{ d = 1 } ^ { D_k } \left[ \ln \Gamma \left( \frac{ N + \nu_k + 1 - d } { 2 } \right) + \ln \Gamma \left( \frac{ \nu_k + 1 - d } { 2 } \right) \right] \right\}. \nonumber
\end{eqnarray}
Another form for $s ( \vect{X}_i, \vect{X}_j )$ is
\begin{equation} \label{eq:s:deltaphi}
 s ( \vect{X}_i, \vect{X}_j ) = \Delta \phi_{ i \cup j } - \Delta \phi_i - \Delta \phi_j,
\end{equation}
with
\begin{equation} \label{eq:deltaphi}
 \Delta \phi_k = \phi ( N + \nu_k, \matr{\Lambda}_k + \matr{S}_k ) - \phi ( \nu_k, \matr{\Lambda}_k ), \qquad k \in \{ i, j, i \cup j \},
\end{equation}
and
\begin{equation} \label{eq:phi}
 \phi ( n, \matr{A} ) = - \frac{ n } { 2 } \ln | \matr{A} | + \sum_{ d = 1 } ^ { \dim ( \matr{A} ) } \ln \Gamma \left( \frac{ n + 1 - d } { 2 } \right). 
\end{equation}
$\Delta \phi_k$ quantifies, up to a constant that cancels out in $s ( \vect{X}_i, \vect{X}_j )$, the amount by which the data support a model of multivariate normal distributions with unrestricted covariance matrix for $\vect{X}_k$.

\subsection{Asymptotic form of the log Bayes factor}

We can now provide an asymptotic expression for $s ( \vect{X}_i, \vect{X}_j )$. Define $\wh{\matr{S}}_k$ as the standard sample covariance matrix, i.e., $\matr{S}_k = N \wh{\matr{S}}_k$. When $N \to \infty$, we can use Stirling approximation \gcite[p.~257]{Abramowitz-1972} to expand the expression $\phi$ of Equation~(\ref{eq:phi}), leading to (see Appendix~\ref{ann:asympt})
\begin{eqnarray} \label{eq:bic}
 s ( \vect{X}_i, \vect{X}_j ) & = & \frac{ N } { 2 } \ln \frac{ | \wh{\matr{S}}_i | \, | \wh{\matr{S}}_j | } { | \wh{\matr{S}}_{ i \cup j } | } - \frac{ 1 } { 2 } \left[ \frac{ D_{ i \cup j } ( D_{ i \cup j } + 1 ) } { 2 } - \sum_{ k \in \{ i, j \} } \frac{ D_k ( D_k + 1 ) } { 2 } \right] \ln N + O ( 1 ) \nonumber \\
 & = & N \, \hat{I} ( \vect{X}_i, \vect{X}_j ) - \frac{ D_i D_j } { 2 } \ln N + O ( 1 ), 
\end{eqnarray}
where
\begin{equation*}
 \hat{I} ( \vect{X}_i, \vect{X}_j ) = \frac{ 1 } { 2 } \ln \frac{ | \wh{\matr{S}}_i | \, | \wh{\matr{S}}_j | } { | \wh{\matr{S}}_{ i \cup j } | }
\end{equation*}
is the plug-in estimator of mutual information for a multivariate normal distribution. Alternatively, $N \, \hat{I} ( \vect{X}_i, \vect{X}_j )$ can be seen as the minimum discrimination information for the independence of $\vect{X}_i$ and $\vect{X}_j$ \gcite[Chap.~12, \S3.6]{Kullback-1968}.

\subsection{Hierarchical agglomerative clustering}

A hierarchy on a set of $D$ variables is a nested set of partitions $( \mathcal{C}_l )_{ l = 0 } ^ D$, where $\mathcal{C}_l$ is a partition of $\{ 1, \dots, D \}$ into $D - l$ clusters \gcite{Duda-2000}. A hierarchical agglomerative clustering (AHC) is a generic procedure to generate such a hierarchy, outlined in pseudo-code in Algorithm~\ref{algo_ahc}. The main steps of the algorithm are: (1) Initialize the partition with singletons (line 2); (2) derive a matrix $\matr{s}_l$ where each element represents the similarity between two clusters $\vect{X}_i$ and $\vect{X}_j$ of $\mathcal{C}_l$, based on an arbitrary function $s ( \vect{X}_i, \vect{X}_j )$ (line 4); (3) identify the two clusters that are most similar\footnote{There may be more than one pair of clusters which maximize the similarity function. Most implementations of AHC proceed by selecting arbitrarily one such pair (e.g., the first one to be detected). In the in-house implementation we used, the pair was selected randomly amongst all these pairs. This was done to properly capture the instability of the algorithm. In such a form, AHC may not be deterministic anymore, in that two runs of the same algorithm on the same dataset may result in different hierarchies.} (line 5); (4) form a new partition identical to the previous one, except that the two most similar clusters are now merged (lines 6--7); (5) iterate Steps~2--4 until the partition has only one single element which covers the whole set of variables (line 3). In the case of the methods proposed here, the similarity measure is given by either Equation~(\ref{eq:bayes:def:2a}) for the exact formulation or Equation~(\ref{eq:bic}) for the asymptotic BIC approximation.

\begin{algorithm}[!htbp]
 \caption{General description of the hierarchial agglomerative clustering algorithm.} \label{algo_ahc}
 \begin{algorithmic}[1]
  \RETURN Hierarchy $(\mathcal{C}_l)_{ l = 0 } ^ D$ 
  \STATE $\mathcal{C}_0 \leftarrow \left\{ \{ X_1 \}, \dots, \{ X_D \} \right\}$
  \FOR{$l = 0, \dots, D - 2$}
  \STATE $\matr{s}_{l} \leftarrow \left[ s ( \vect{X}_i, \vect{X}_j ) \right]_{ \vect{X}_i, \vect{X}_j \in \mathcal{C}_l }$
  \STATE $( i^*, j^* ) \leftarrow \operatorname*{arg\,max}_{ i, j } \matr{s}_l ( i, j )$
  \STATE $\mathcal{C}_{l+1} \leftarrow \mathcal{C}_l \setminus \{ i^*,j^*\}$
  \STATE $\mathcal{C}_{l+1} \leftarrow \mathcal{C}_l \cup \{ i^*\cup j^*\}$
 \ENDFOR
 \end{algorithmic}
\end{algorithm}

\subsection{Comparing distinct levels of the hierarchy}

The last development aims at providing a way to compare nested partitions, i.e., different levels of the hierarchy. Once the hierarchical clustering has been performed, each step is associated with a partition of $\vect{X}$. Assume that, at level $l$, the partition reads $\{ \vect{X}_1, \dots, \vect{X}_K \}$ and that, at step $l+1$, $\vect{X}_i$ and $\vect{X}_j$ are clustered. Denote by ${\mathcal C}_l$ the assumption that the partition at level $l$ is the correct partition of $\vect{X}$. The global improvement brought by the clustering of $\vect{X}_i$ and $\vect{X}_j$ between steps $l$ and $l + 1$ can be quantified by the log ratio between the marginal model likelihoods
$$\ln \frac{ \pr ( \matr{S} | {\mathcal C}_{ l + 1 } ) } { \pr ( \matr{S} | {\mathcal C}_l ) },$$
where both quantities can be computed in a manner similar to what was done for the similarity measure. For instance, if ${\mathcal C}_l$ is true, then $\matr{\Sigma}$ is block-diagonal with $K$ blocks $\matr{\Sigma}_k$'s, the submatrix restrictions of $\matr{\Sigma}$ to $\vect{X}_k$. Introducing the model parameters then yields 
\begin{equation} \label{eq:part:margth}
 \pr ( \matr{S} | {\mathcal C}_l ) = \int \pr ( \matr{S} | {\mathcal C}_l, \matr{\Sigma}_1, \dots, \matr{\Sigma}_K ) \, \pr ( \matr{\Sigma}_1, \dots, \matr{\Sigma}_K | {\mathcal C}_l ) \prod_{ k = 1 } ^ K \ud \matr{\Sigma}_k.
\end{equation}
In a way similar to what was done previously, the likelihood $\pr ( \matr{S} | {\mathcal C}_l, \matr{\Sigma}_1, \dots, \matr{\Sigma}_K )$ can be expanded as
\begin{equation} \label{eq:part:vrais}
 \pr ( \matr{S} | {\mathcal C}_l, \matr{\Sigma}_1, \dots, \matr{\Sigma}_K ) = \frac{ | \matr{S} | ^ { \frac{ N - D - 1 } { 2 } } } { Z ( D, N ) } \prod_{ k = 1 } ^ K | \matr{\Sigma}_k | ^ { - \frac{ N } { 2 } } \exp \left[ - \frac{ 1 } { 2 } \tr ( \matr{\Sigma}_k ^ { - 1 } \matr{S}_k ) \right].
\end{equation}
Turning to the prior distribution $\pr ( \matr{\Sigma}_1, \dots, \matr{\Sigma}_K | {\mathcal C}_l )$ and assuming no prior dependence between the $\matr{\Sigma}_k$'s, we can set
\begin{equation} \label{eq:part:prior:1}
 \pr ( \matr{\Sigma}_1, \dots, \matr{\Sigma}_K | {\mathcal C}_l ) = \prod_{ k = 1 } ^ K \pr ( \matr{\Sigma}_k | {\mathcal C}_l ).
\end{equation}
Each $\pr ( \matr{\Sigma}_k | {\mathcal C}_l )$ can be obtained by the marginalization of $\pr ( \matr{\Sigma} | {\mathcal C}_l )$, which is here taken as an inverse-Wishart distribution with $\nu$ degrees of freedom and inverse scale matrix the $D$-by-$D$ diagonal matrix $\matr{\Lambda}$. Note that such a prior on $\matr{\Sigma}$ is compatible with the prior used earlier for $\matr{\Sigma}_{ i \cup j }$ if one sets $\nu_{ i \cup j } = \nu - D + D_{ i \cup j }$ and if $\matr{\Lambda}_{ i \cup j }$ is the restriction of $\matr{\Lambda}$ to $\vect{X}_{ i \cup j }$ \gcite[\S5.2]{Press_SJ-2005}. We then have
\begin{equation} \label{eq:part:prior:2}
 \pr ( \matr{\Sigma}_k | {\mathcal C}_l ) = \frac{ | \matr{\Lambda}_k | ^ { - \frac{ \nu_k }{ 2 } } } { Z ( D_k, \nu_k ) } | \matr{\Sigma}_k | ^ { - \frac{ \nu_k + D_k + 1 }{ 2 } } \exp \left[ - \frac{ 1 } { 2 } \tr ( \matr{\Lambda}_k \matr{\Sigma}_k ^ { - 1 } ) \right].
\end{equation}
Incorporating Equations (\ref{eq:part:vrais}), (\ref{eq:part:prior:1}), and (\ref{eq:part:prior:2}) into Equation~(\ref{eq:part:margth}) and integrating leads to
\begin{equation} \label{eq:part:marg}
 \pr ( \matr{S} | {\mathcal C}_l ) = \frac{ | \matr{S} | ^ { \frac{ N - D - 1 } { 2 } } } { Z ( D, N ) } \prod_{ k = 1 } ^ K \frac{ Z ( D_k, N + \nu_k ) }{ Z ( D_k, \nu_k ) } \frac{ | \matr{\Lambda}_k | ^ { \frac{ \nu_k } { 2 } } } { | \matr{\Lambda}_k + \matr{S}_k | ^ { \frac{ N + \nu_k } { 2 } } }.
\end{equation}
The same calculation can be performed for $\pr ( \matr{S} | {\mathcal C}_{ l + 1 } )$. The result is the same as in Equation~(\ref{eq:part:marg}), except that the product is composed of $K - 1$ terms. Of these terms, $K - 2$ correspond to clusters that are unchanged from ${\mathcal C}_l$ to ${\mathcal C}_{ l + 1 }$ and, as a consequence, are identical to those of Equation~(\ref{eq:part:marg}). The $( K - 1)$th term corresponds to the cluster obtained by the merging of $\vect{X}_i$ and $\vect{X}_j$. As a consequence, the log Bayes factor reads
\begin{eqnarray*}
 \ln \frac{ \pr ( \matr{S} | {\mathcal C}_{ l + 1 } ) } { \pr ( \matr{S} | {\mathcal C}_l ) } & = & \ln \left[ \frac{ Z ( D_{ i \cup j }, N + \nu_{ i \cup j } ) }{ Z ( D_{ i \cup j }, \nu_{ i \cup j } ) } \frac{ | \matr{\Lambda}_{ i \cup j } | ^ { \frac{ \nu_{ i \cup j } } { 2 } } } { | \matr{\Lambda}_{ i \cup j } + \matr{S}_{ i \cup j } | ^ { \frac{ N + \nu_{ i \cup j } } { 2 } } } \right] \\
 & &  \qquad - \sum_{ k \in \{ i, j \} } \ln \left[ \frac{ Z ( D_k, N + \nu_k ) } { Z ( D_k, \nu_k ) } \frac{ | \matr{\Lambda}_k | ^ { \frac{ \nu_k } { 2 } } } { | \matr{\Lambda}_k + \matr{S}_k | ^ { \frac{ N + \nu_k } { 2 } } } \right].
\end{eqnarray*}
But this quantity is nothing else than $s ( \vect{X}_i, \vect{X}_j )$. In other words, we proved that
\begin{equation} \label{eq:global:local}
 \ln \frac{ \pr ( \matr{S} | {\mathcal C}_{ l + 1 } ) } { \pr ( \matr{S} | {\mathcal C}_l ) } = s ( \vect{X}_i, \vect{X}_j ),
\end{equation}
i.e., $s ( \vect{X}_i, \vect{X}_j )$, the \emph{local} measure of similarity between $\vect{X}_i$ and $\vect{X}_j$, can be used to compute the \emph{global} measure of relative probability between two successive levels of the hierarchy.

\subsection{Setting the hyperparameters}

Hyperparameter selection is often a thorny issue in Bayesian analysis. We here considered two approaches. The first approach (coined \texttt{BayesCov}) is to set the degree of freedom to the lowest value that still corresponds to a well-defined distribution, that is $\nu = D$, and a diagonal scale matrix that optimizes the marginal model likelihood of Equation~(\ref{eq:part:marg}) before any clustering (that is, with $K = D$ clusters and $D_k = 1$ for all $k$), yielding (see Appendix~\ref{ann:optim})
\begin{equation*}
 \Lambda_{ d d } = \frac{ \nu - D + 1 } { N } S_{ d d },
\end{equation*}
where $\Lambda_{dd}$ and $S_{dd}$ are the diagonal elements of the prior inverse scale matrix $\matr{\Lambda}$ and sum-of-square matrix $\matr{S}$, respectively. An alternative approach (coined \texttt{BayesCorr}) is to work with the sample correlation matrix instead of the sample covariance matrix. One can then set the number of degrees of freedom to $\nu = D + 1$ and the scale matrix to the identity matrix. The corresponding prior distribution yields uniform marginal distributions for the correlation coefficients \gcite{Barnard_J-2000}. Note that clustering with the asymptotic form of Equation~(\ref{eq:bic}) (coined \texttt{Bic}) does not involve hyperparameters; it is also insensitive to the fact that the input is the covariance matrix or the correlation matrix.

\subsection{Automatic stopping rule}

An advantage of the Bayesian clustering scheme proposed here and its BIC approximation is that they come naturally with an automatic stopping rule. By definition of $s$ in Equation~(\ref{eq:s}), the fact that $s ( \vect{X}_{ i_0 }, \vect{X}_{ j_0 } ) > 0$ for the pair that is selected for clustering also means that the marginal model likelihood for ${\cal M}_D$ is larger than that for ${\cal M}_I$. As such, $\vect{X}_{ i_0 }$ and $\vect{X}_{ j_0 }$ are more likely to belong to the same cluster than not and, as a consequence, it indeed makes sense to cluster them. By contrast, if we have $s ( \vect{X}_{ i_0 }, \vect{X}_{ j_0} ) < 0$ for the same pair, the marginal model likelihood for ${\cal M}_D$ is smaller than that for ${\cal M}_I$. $\vect{X}_{ i_0 }$ and $\vect{X}_{ j_0 }$ are therefore more likely to belong to different clusters. If, at a given step of the clustering, the pair with highest similarity measure has a negative similarity measures, then all pairs do, meaning that all pairs of
clusters tested more probably belong to different clusters. It therefore makes sense to stop the clustering procedure at that point. This shows that an automatic stopping rule can simply be implemented into the clustering algorithm: Stop the clustering if the pair $( \vect{X}_{ i_0 }, \vect{X}_{ j_0} )$ selected for clustering at a given step has $s ( \vect{X}_{ i_0 }, \vect{X}_{ j_0 } ) < 0$. Note that, according to Equation~(\ref{eq:global:local}), the resulting clustering corresponds to the one in the hierarchy that has largest marginal likelihood. We will refer to \texttt{BayesCovAuto}, \texttt{BayesCorrAuto} and \texttt{BicAuto} when applying the clustering schemes with this automated stopping rule.

\section{Results}

\subsection{Validation on synthetic data}

To assess the behavior of the method expounded here, we examined how it fared on synthetic data. We used the two variants of the Bayes factor mentioned above (\texttt{BayesCov} and \texttt{BayesCorr}), \texttt{Bic}, as well as the same methods with automatic stopping rule (\texttt{BayesCovAuto}, \texttt{BayesCorrAuto} and \texttt{BicAuto}). As a mean of comparison, we also used the following methods---
\begin{itemize}
 \item  A random hierarchical clustering scheme, where variables were clustered uniformly at random at each step. This category contains only one algorithm: \texttt{Random}, which was implemented for the purpose of the present study.
 \item Hierarchical clustering schemes with similarity measures given by either Pearson correlation or absolute Pearson correlation, and a merging rule based on either the single, average, or complete linkage, or using Ward criterion. This category contains 8 algorithms: \texttt{Single}, \texttt{Average}, \texttt{Complete}, \texttt{Ward}, \texttt{SingleAbs},  \texttt{AverageAbs}, \texttt{CompleteAbs}, \texttt{WardAbs}. We used the implementations of these methods proposed in NIAK\footnote{\texttt{https://github.com/SIMEXP/niak}}
 \item Hierarchical clustering with a similarity measure given by mutual information, with and
without normalization. This category contains 2 algorithms: \texttt{Infomut} and \texttt{InfomutNorm}. These methods were implemented for the purpose of the present study. Note that neither algorithm can run on small samples.
 \item An approach where the clusters are estimated as the blocks of the precision (i.e., inverse covariance) matrix estimated with the graphical lasso---essentially a maximum-likelihood with $L_1$-norm penalization \gcite{Friedman_J-2008}. The penalization parameter $\lambda$ was set in $[0, 1]$ by step of 0.1, then to 5, 10, 20, 50, 100, 200 \gcite{Smith_SM-2011}. A version that optimizes $\lambda$ with a BIC criterion was also used \gcite{Lian-2011}. Since the graphical lasso is not invariant by transformation of a covariance matrix into a correlation matrix, we used either the covariance matrix or the correlation matrix as input. Note that this approach automatically determined a number of clusters. Also, for $\lambda = 0$ (unconstrained case), the algorithm cannot run on small samples. This category contains 34 algorithms: 16 algorithms \texttt{gLassoCov-x} and 16 algorithms \texttt{gLassoCorr-x}, where \texttt{x} is the value of $\lambda$, and 2 algorithms \texttt{gLassoCovOpt} and \texttt{gLassoCorrOpt}. For this category of algorithms, we used a package freely available\footnote{\texttt{http://www.cs.ubc.ca/$\sim$schmidtm/Software/L1precision.html}} and already used in \gcite{Smith_SM-2011}.
 \item An approach based on the spectral clustering \gcite{von_Luxburg-2006} of either the raw value or the absolute value of either the correlation or the partial correlation matrix. This approach required the number of clusters as input. Since this clustering has a step of $k$-means, which is stochastic by nature, we considered 2 variants: one with 1 step of $k$-means and the other one with 10 repetitions of $k$-means and selection of the best clustering in terms of inertia. The similarity measures were defined so that the range would be the same (namely in $[ 0, 1 ]$) when using the signed or the absolute value of correlation: $0.5 ( 1 + r )$ and $| r |$, respectively. This category contains 8 algorithms:  2 algorithms \texttt{SpecCorr-x}, 2 algorithms \texttt{SpecCorrAbs-x}, 2 algorithms \texttt{SpecCorrpar-x} and 2 algorithms \texttt{SpecCorrparAbs-x}, where \texttt{x} is the number of times that $k$-means is performed. The spectral clustering algorithms of this category were coded for the purpose of the present study, while we used the implementations of the $k$-means algorithm proposed in NIAK.
\end{itemize}
All in all, 59 variants were tested.

\paragraph{Data description.}

In order to assess the performance of the Bayesian approach, we performed the following set of simulations. For each value of $D$ in $\{6, 10, 20, 40 \}$, we considered partitions with increasing number of clusters $C$ ($1 \leq C \leq D$). For a given value of $C$, we performed 500 simulations. For each simulation, the $D$ variables were randomly partitioned into $C$ clusters, all partitions having equal probability of occurrence \gcite[chap.~12]{Nijenhuis-1978}; \gcite{Wilf-1999}. For a given partition $\{ \vect{X}_1, \dots, \vect{X}_K \}$ of $\vect{X}$, we generated data according to 
\begin{equation} \label{eq:simu:mod}
 f ( \vect{x} ) = \prod_{ k = 1 } ^ K f_k ( \vect{x}_k ),
\end{equation}
where all $f_k$'s were taken either as multivariate normal distributions (parameters: mean $\vect{\mu}_k$ and covariance matrix $\matr{\Sigma}_k$) or multivariate Student-$t$ distributions (parameters: degres of freedom $\nu$, location parameter $\vect{\mu}_k$ and scale matrix $\matr{\Sigma}_k$). In both case, the $\vect{\mu}_k$'s were set to $\vect{0}$ and the $\matr{\Sigma}_k$'s were first sampled according to a Wishart distribution with $D_k + 1$ degrees of freedom and scale matrix the identity matrix and then rescaled to a correlation matrix. The sampling scheme on $\matr{\Sigma}_k$ generated correlation matrices with uniform marginal distributions for all
correlation coefficients \gcite{Barnard_J-2000}. For the multivariate Student-$t$ distributions, $\nu$ was set in $\{ 1, 3, 5 \}$. Equation~(\ref{eq:simu:mod}) was used to generate synthetic datasets of length $N$ varying from 10 to 300 by increment of 40. Each dataset was summarized by its sample correlation matrix and hierarchical clustering was performed using the methods mentioned above. All simulations were implemented using the Pipeline System for Octave and Matlab, PSOM\footnote{\texttt{https://github.com/SIMEXP/psom}} \gcite{Bellec-2012} under Matlab~7.2 (The MathWorks, Inc.) and run on a 24-core server.
\par
To assess the efficiency of the various methods, we thresholded each clustering at the true number of clusters, except for \texttt{BayesCovAuto}, \texttt{BayesCorrAuto}, \texttt{BicAuto} and \texttt{gLasso} for which we used the clustering obtained by the method. We then quantified the quality of the resulting partition using the proportion of correct classifications as well as the adjusted Rand index, which computes the fraction of variable pairs that are correctly clustered corrected for chance \gcite{Rand-1971, Hubert-1985}. Results corresponding to a given dimension $D$ and a given method were then pooled across numbers of clusters $C$, lengths $N$ and distributions (multivariate normal and Student). We classified the methods from best to worst based on these global results using the following indices (in this order): median of adjusted Rand index, 25\% percentile value of adjusted Rand index, 5\% percentile value of adjusted Rand index, smallest value of adjusted Rand index, and proportion of correct classifications. Note that some algorithms (\texttt{Infomut}, \texttt{InfomutNorm} and \texttt{SpecCorrpar}) require the sample covariance matrix to be definite positive. As a consequence, these algorithms could not run on small samples. We therefore restrained our evaluation to cases where all algorithms were operational. Finally, we performed a Bayesian ANOVA-like regression analysis \gcite[\S15.6]{Gelman-2004b}, where we explained the adjusted Rand index of nine algorithms (\texttt{BayesCov}, \texttt{BayesCovAuto}, \texttt{BayesCorr}, \texttt{BayesCorrAuto}, \texttt{Bic}, \texttt{BicAuto}, \texttt{Infomut}, \texttt{InfomutNorm}, and \texttt{AverageAbs}) with the following effects: clustering algorithm (9 levels), number of variables $D$ (4 levels: $D \in \{ 6, 10, 20, 40 \}$), type of distribution (4 levels: multivariate Gaussian or multivariate Student-$t$ with 1, 3, or 5 degrees of freedom), number of samples $N$ (8 levels: $N \in \{ 10, 50, 90, 130, 170, 210, 250, 290 \}$), and number of clusters $C$ (68 levels: from 2 to $D - 1$ clusters for each $D$). In other words, the model considered was of the form
\begin{equation} \label{eq:simu:regr}
 \mbox{adj Rand index} ( \mathrm{algo}, D, \mathrm{distr}, N, C ) = \beta_0 + \beta_{ \mathrm{algo} } + \beta_D + \beta_{ \mathrm{distr} } + \beta_N + \beta_{ D, C } + \epsilon.
\end{equation}
Interactions between effects, while potentially relevant, were not considered to keep the analysis tractable. The posterior distribution for the various regression parameters were estimated using Gibbs sampling.

\paragraph{Results.}

\begin{figure}[!htbp]
 \centering
 \includegraphics[width=8cm]{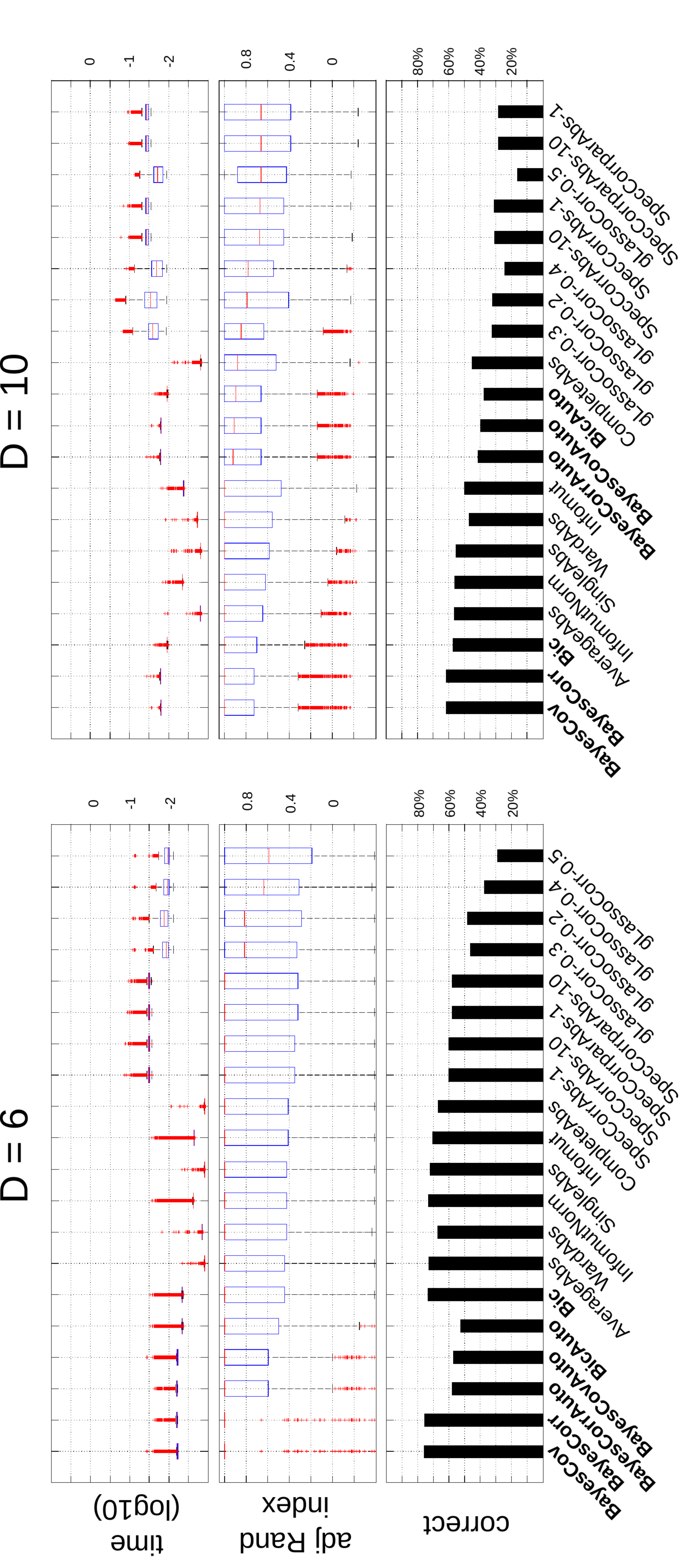}
 \caption{{\bf Simulation study.} Computational time (top), adjusted Rand index (middle) and proportion of correct classifications (bottom) for $D = 6$ (left) and $D = 10$ (right).} \label{fig:simu:1}
\end{figure}

\begin{figure}[!htbp]
 \centering
 \includegraphics[width=8cm]{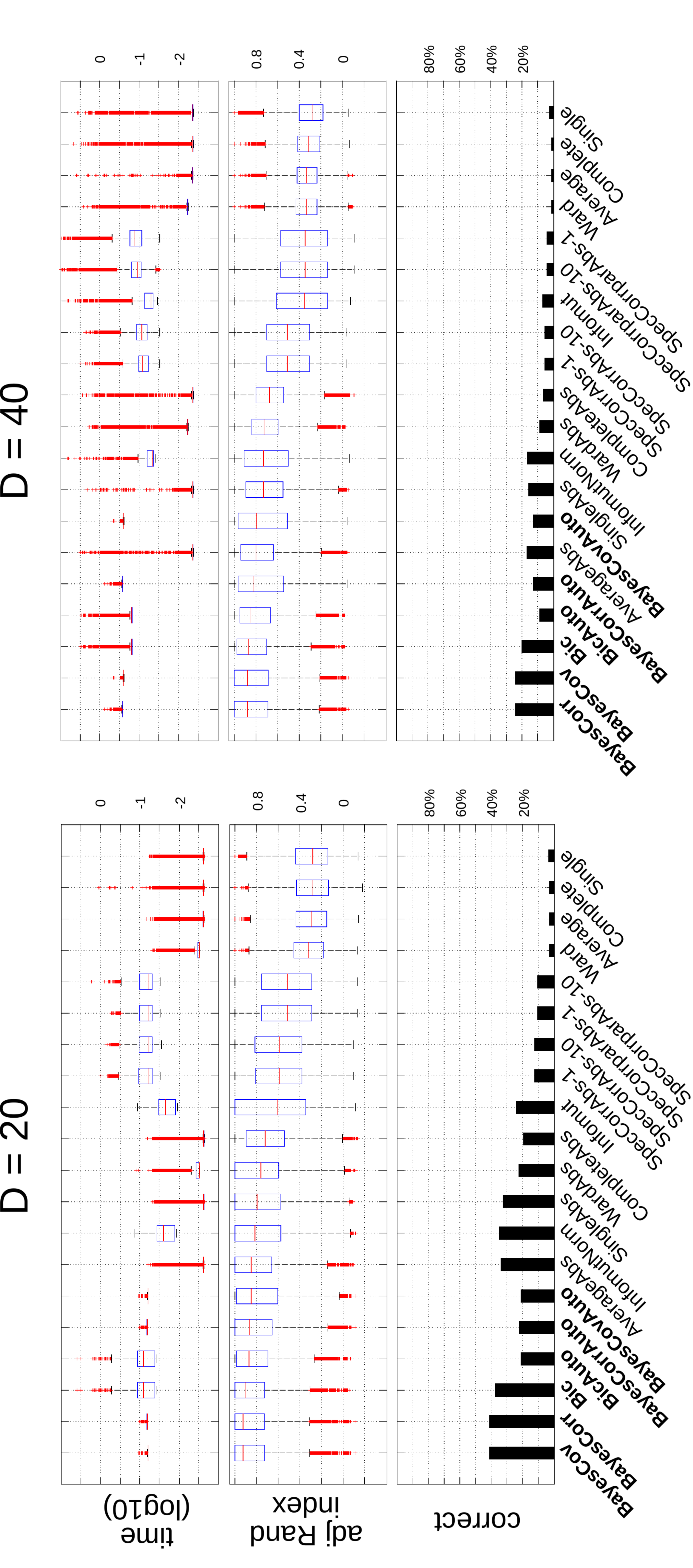}
 \caption{{\bf Simulation study.} Computational time (top), adjusted Rand index (middle) and proportion of correct classifications (bottom) for $D = 20$ (left) and $D = 40$ (right).} \label{fig:simu:2}
\end{figure}

The results corresponding to the adjusted Rand index and fraction of correct classification are summarized in Figs \ref{fig:simu:1} and \ref{fig:simu:2} for the 20 best methods. Globally, and as expected, all methods were adversely affected by an increase in the number of variables. In all cases, the variants proposed in the present paper performed very well compared to other methods. \texttt{BayesCov} and \texttt{BayesCorr} were always classified as the two best algorithms and \texttt{Bic} was never outperformed by a method already published. The methods with automatic thresholding of the hierarchy performed surprinsingly well, considered that they were compared against methods with oracle. In particular, they clearly outperformed all variants of \texttt{gLasso}, the only method that was able to automatically determine the number of clusters. Of note, all variants of \texttt{gLasso} proved too slow to be applied to our simulation data for $D \in \{ 20, 40 \}$.

\begin{figure}[!htbp]
 \centering
 \includegraphics[width=10cm]{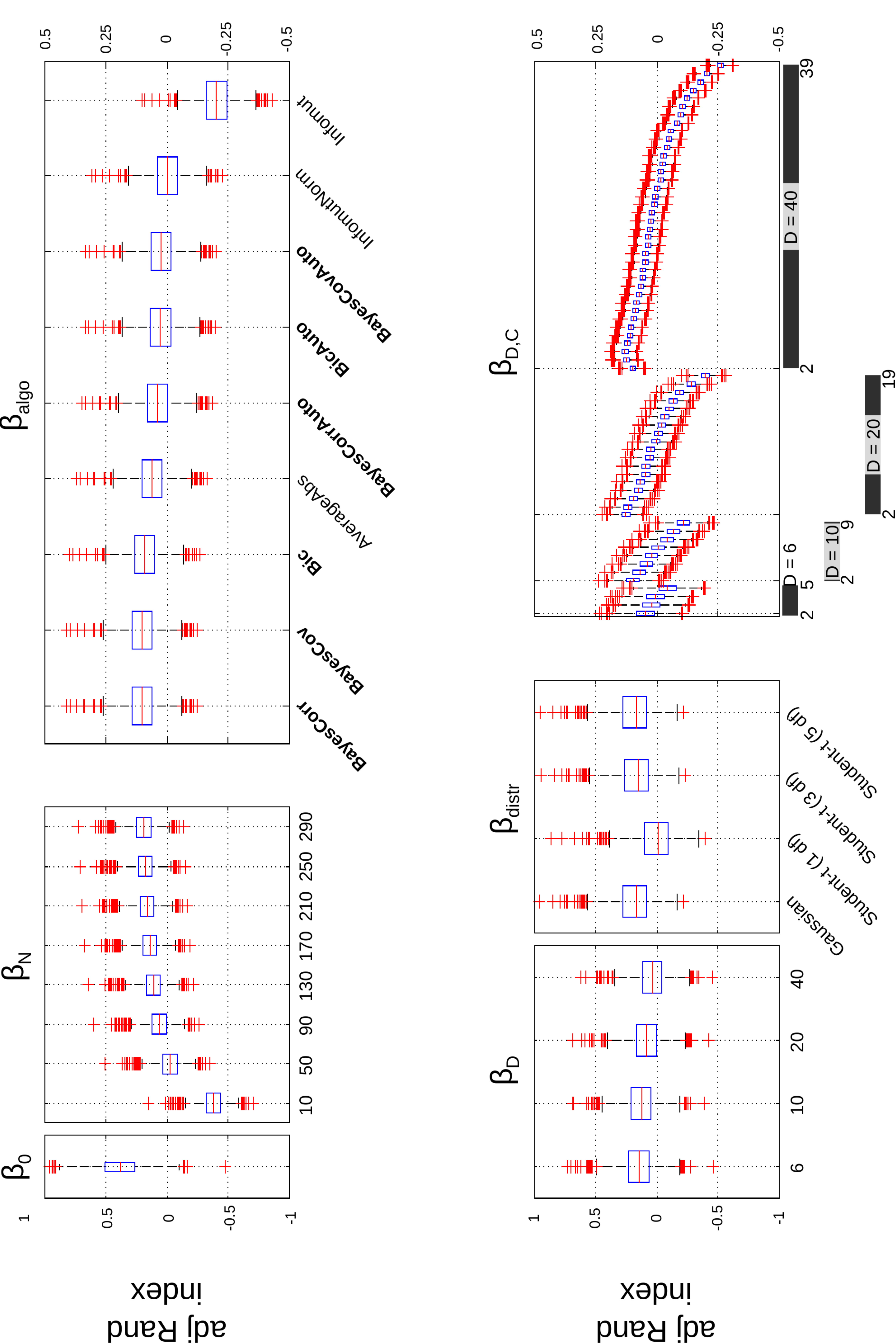}
 \caption{{\bf Simulation study.} Result of the regression analysis. Posterior distribution for the different regression coefficients: $\beta_0$ (global effect), $\beta_N$ (dataset size $N$), $\beta_{ \mathrm{algo} }$ (algorithm), $\beta_D$ (number of variables $D$), $\beta_{ \mathrm{distr} }$ (type of distribution), and $\beta_{ D, C }$ (number of clusters) [see Equation~(\ref{eq:simu:regr})].} \label{fig:simu:3}
\end{figure}

The results of the regression analysis are represented in Fig~\ref{fig:simu:3}. The 9 algorithms selected included the ones proposed in the present manuscript (\texttt{BayesCov}, \texttt{BayesCovAuto}, \texttt{BayesCorr}, \texttt{BayesCorrAuto}, \texttt{Bic}, and \texttt{BicAuto}),  the best-performing classic algorithm in the previous analysis (\texttt{AverageAbs}) as wel as the algorithms based on mutual information (\texttt{Infomut} and \texttt{InfomutNorm}). We found that increasing the dataset size ($N$) increased the performance of the algorithm, while increasing the dimensionality of the problem ($D$) and the number of clusters ($C$) decreased it. Note that dimension was found to have a negative influence on the adjusted Rand index, even though this index was partly proposed as a modification from the raw Rand index to overcome this limitation. Finally, this analysis confirmed the superior behavior of the methods proposed here, even when the method included the automatic stopping rule.

\subsection*{Toy example}

This data set was first used in \cite{Roverato-1999} and later re-analyzed in \cite{Marrelec-2006b} in the context of conditional independence graphs. It originates from a study investigating early diagnosis of HIV infection in children from HIV positive mothers. The variables are related to various measures on blood and its components: $X_1$ and $X_2$ immunoglobin G and A, respectively; $X_4$ the platelet count; $X_3$, $X_5$ lymphocyte B and T4, respectively; and $X_6$ the T4/T8 lymphocyte ratio. The sample summary statistics are given in Table~\ref{tab:ex:donnees}. \cite{Roverato-1999} found that the correlations between $X_4$ and any other variable had strong probability around zero and hypothesized that the model was overparametrized. Based on the simulation study, we performed clustering of the data with the following methods: \texttt{BayesCov}(\texttt{Auto}), \texttt{BayesCorr}(\texttt{Auto}), \texttt{Bic}(\texttt{Auto}), \texttt{Infomut}, \texttt{InfomutNorm}, \texttt{SingleAbs}, \texttt{AverageAbs}, \texttt{CompleteAbs}, \texttt{WardAbs}, \texttt{SpecCorrAbs} and \texttt{SpecCorrparAbs}. For spectral clustering, we used either 1 or 10 repetitions of the $k$-means step; since the results obtained for 1 step of $k$-means were highly variable for 3, 4, and 5 clusters, we discarded these results.

\begin{table}[!htbp]
 \centering
 \caption{{\bf Toy example.} Summary statistics for the HIV data: sample variances (main diagonal), correlations (lower triangle) and partial correlations (upper triangle) [from \cite{Roverato-1999}].}
\label{tab:ex:donnees}
 \begin{tabular}{c|cccccc}
  $x_1$ & 8.8374 & 0.479 & $- 0.043$ & $-0.033$ & 0.356 & $-0.236$ \\
  $x_2$ & 0.483 & 0.1919 & 0.068 & $-0.084$ & $-0.224$ & $-0.110$ \\
  $x_3$ & 0.220 & 0.057 & 8924231.9 & 0.085 & 0.552 & $-0.330$ \\
  $x_4$ & $-0.040$ & $-0.133$ & 0.149 & 20392.4 & 0.091 & 0.013 \\
  $x_5$ & 0.253 & $-0.124$ & 0.523 & 0.179 & 1952795.2 & 0.384 \\
  $x_6$ & $-0.276$ & $-0.314$ & $-0.183$ & 0.064 & 0.213 & 1.378 \\
  \hline 
  & $x_1$ & $x_2$ & $x_3$ & $x_4$ & $x_5$ & $x_6$
 \end{tabular}
\end{table}

\begin{figure}[!htbp]
 \centering
 \includegraphics[width=10cm]{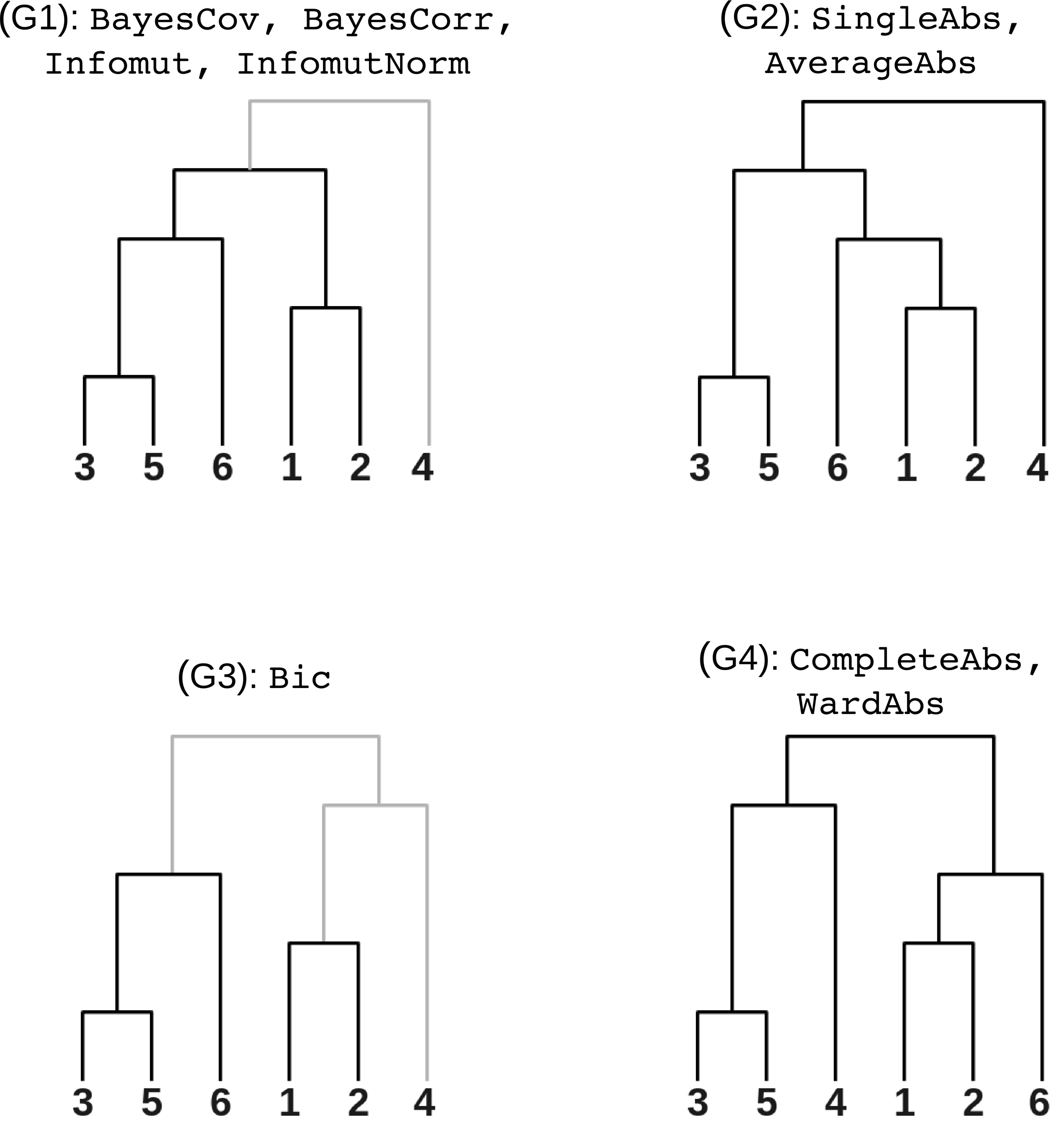} 
 \caption{{\bf Toy example.} Result of clustering. Algorithms in the top row clustered $X_4$ at the last step, while it was clustered at the before the last step for algorithms in the bottom row. Algorithms in the left column clustered $X_6$ with $\{ X_3, X_5 \}$, while $X_6$ was clustered with $\{ X_1, X_2 \}$ for the algorithms in the right column. Parts in grey correspond to clustering steps that were not performed by \texttt{BayesCovAuto} or \texttt{BayesCorrAuto} in (G1), or \texttt{Bic} in
(G3).}
\label{fig:ex:res:1}
\end{figure}

The resulting clusterings are given in Fig~\ref{fig:ex:res:1} and Table~\ref{tab:ex:res}. All hierarchical clusterings started by clustering $X_3$ and $X_5$ (lymphocite). This was confirmed by \texttt{SpecCorrAbs-10} when required to provide 5 clusters. All hierarchical clustering methods then clustered $X_1$ and $X_2$ (immunoglobin). This result was in agreement with both \texttt{SpecCorrAbs-10} and \texttt{SpecCorrparAbs-10} when required to provide 4 clusters. After the second step, we observed four behaviors for the AHC algorithms and classified them accordingly:
\begin{itemize}
 \item[(G1)] \texttt{BayesCov}, \texttt{BayesCorr}, \texttt{Infomut} and
\texttt{InfomutNorm};
 \item[(G2)] \texttt{SingleAbs} and \texttt{AverageAbs};
 \item[(G3)] \texttt{Bic};
 \item[(G4)] \texttt{CompleteAbs} and \texttt{WardAbs}.
\end{itemize}
While not a hierarchical clustering, \texttt{SpecCorrAbs-10} provided successive clusterings that were in agreement with methods in (G2). Algorithms in (G1) and (G3) clustered $X_6$ with $\{ X_3, X_5 \}$, creating a cluster of variables related to lymphocite. Algorithms in (G1) and (G2) (and \texttt{SpecCorrAbs-10}) agreed that a partitioning of the variables in two clusters should leads to $\{ X_1, X_2, X_3, X_5, X_6 \}$ on the one hand and $\{ X_4 \}$ on the other hand. This clustering was also found by \texttt{SpecCorrpar-10} when constrained to extract two clusters from the data. It was also considered the best clustering for \texttt{BayesCov} and \texttt{BayesCorr}. For \texttt{Bic}, the optimal partitioning was composed of three clusters, namely, $\{ X_1, X_3, X_6 \}$, $\{ X_1, X_2 \}$, and $\{ X_4 \}$, which is in agreement with what would methods in (G1) yield for three clusters; furthermore, it still kept $X_4$ separated from the other variables.
\par
\begin{figure}[!htbp]
 \centering
 \includegraphics[width=\linewidth]{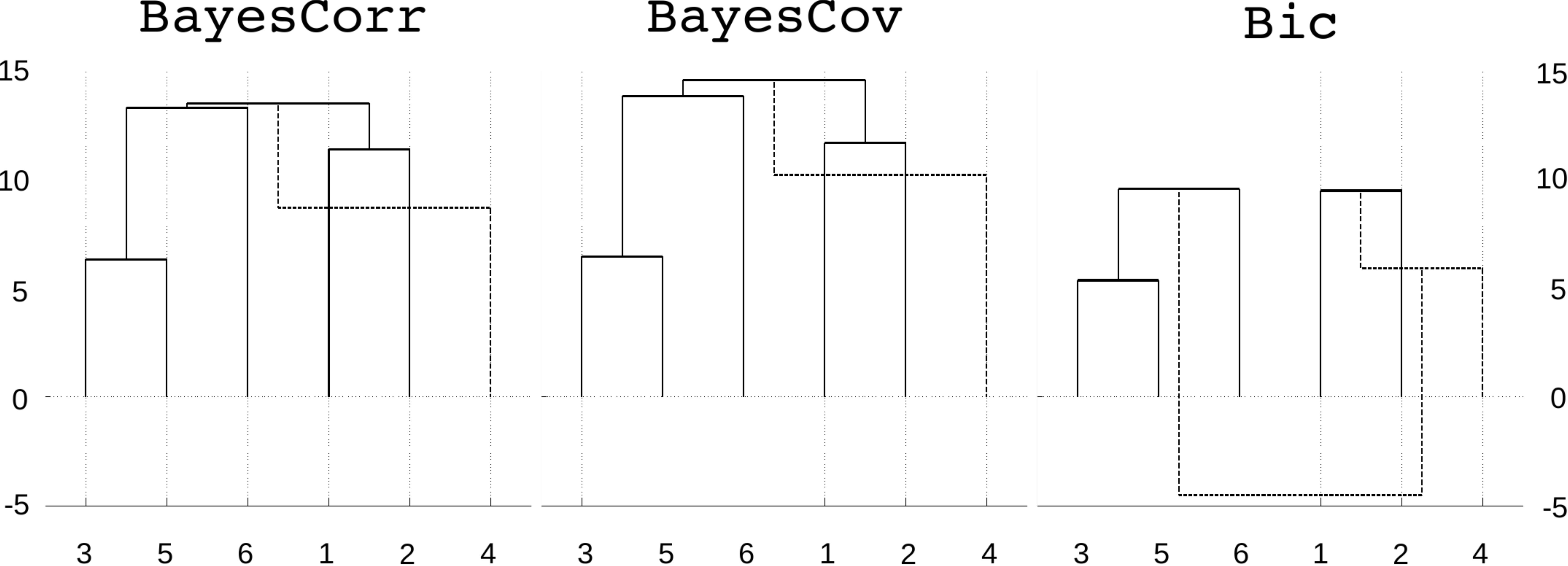}
 \caption{{\bf Toy example.} Result of clustering for \texttt{BayesCorr}, \texttt{BayesCov} and \texttt{Bic}. The values on the $y$ axis correspond to the $\log_{10}$ Bayes factor in favor of the global clustering obtained at each step compared to a model where all variables are independent (step 0 of hierarchical clustering). The dotted lines correspond to clustering steps that were not performed with the automatic stopping rule.}
\label{fig:ex:res:2}
\end{figure}

In Fig~\ref{fig:ex:res:2}, we represented the evolution of the $\log_{10}$ Bayes factor during hierarchical clustering for \texttt{BayesCov}, \texttt{BayesCorr} and \texttt{Bic}. Note that, while the clustering steps are identical for \texttt{BayesCov} and \texttt{BayesCorr}, the log Bayes factors
are similar but not identical. Likewise, while the first two clustering steps of \texttt{Bic} is identical to those of \texttt{BayesCov} and \texttt{BayesCorr}, one can see that, unlike \texttt{BayesCov} and \texttt{BayesCorr}, \texttt{Bic} considered the merging of $\{ X_3, X_5 \}$ with $X_6$ almost as likely as that of $X_1$ with $X_2$. Also, while the successive clusterings of $X_3$ with $X_5$ and then $X_6$ as well as that of $X_1$ with $X_2$ strongly increased the Bayes factor for \texttt{BayesCov} and \texttt{BayesCorr}, the improvement brought by the clustering of $\{ X_3, X_5, X_6 \}$ with $\{ X_1, X_2 \}$ in these methods was less important.
\par
All in all, this analysis led us to the following conclusion: it is very likely that variables $X_1$ and $X_2$ belong to the same cluster of dependent variables, and similarly for variables $X_3$ and $X_5$. Also, there is very strong evidence in favor of the fact that $X_4$ is independent from the other variables. Finally, we suspect that $X_3$, $X_5$ and $X_6$ could belong to the same cluster of variables.

\begin{table}[!htbp]
 \caption{{\bf Toy example.} Result of spectral clustering with increasing number of clusters.} \label{tab:ex:res}
 \centering
 \begin{tabular}{c|cc}
  & \texttt{SpecCorrAbs-10} & \texttt{SpecCorrparAbs-10} \\
  \hline
  2 clusters & \multicolumn{2}{c}{$12356 \mid 4$} \\
  3 clusters & $126 \mid 35 \mid 4$ & $12 \mid 356 \mid 4$ \\
  4 clusters & \multicolumn{2}{c}{$12 \mid 35 \mid 4 \mid 6$} \\
  5 clusters & $1 \mid 2 \mid 35 \mid 4 \mid 6$ & not reproducible
 \end{tabular}
\end{table}

\subsection{fMRI data}

\begin{figure}[!htbp]
 \centering
 \includegraphics[width=\linewidth]{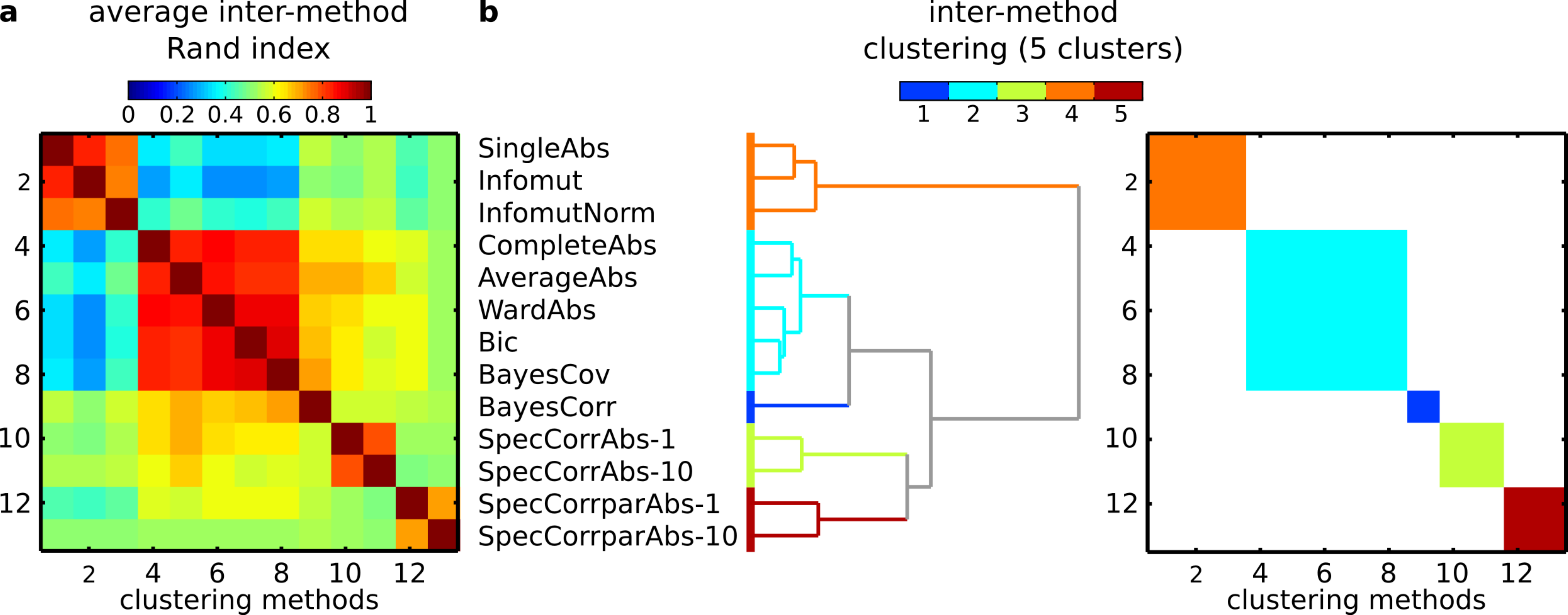}
 \caption{{\bf Real resting-state fMRI data---between-method similarity.} Panel \textbf{a}: Rand indices between individual partitions generated with different methods, averaged across all subjects and scales (number of clusters). Panel \textbf{b}: Hierarchical clustering using matrix shown in Panel~a as a similarity measure and Ward's criterion.}
\label{fig_intermeth}
\end{figure}

Cluster analysis is a popular tool to study the organization of brain networks in resting-state fMRI \gcite{Yeo-2011, Kelly_C-2012}, by identifying clusters of brain regions with highly correlated spontaneous activity. We applied the 13 methods that were found to have good performance on simulations (see Fig~\ref{fig_intermeth}) to a collection of resting-state time series. The time series had 205 time samples and were recorded for 82 brain regions in 19 young healthy subjects. See Appendix~\ref{ann:irmf} for details on data collection and preparation. The data are available online\footnote{\texttt{http://figshare.com/articles/Atlanta\_resting\_state\_fMRI\_time\_series\\ \_preprocessed\_using\_the\_AAL\_template/1521155}}.
\par
We first aimed at establishing which clustering algorithms yielded similar results on these datasets. We more specifically investigated a 7-cluster solution, as this level of decomposition has been examined several times in the literature \gcite{Bellec-2010, Power-2011, Yeo-2011}. Each clustering algorithm was applied to the time series of each subject independently. For a given pair of methods, the similarity between the cluster solutions generated by both methods on the same subject was evaluated with the Rand. Note that the raw, unnormalized Rand index was used here, as we did not compare cluster solutions with different numbers of clusters, which is the main motivation of the normalization. The unnormalized Rand index has a more intuitive interpretation than its adjusted counterpart. The Rand indices were averaged across subjects, hence resulting into a method-by-method matrix capturing the (average) similarity of clustering outputs for each pair of methods (see Fig~\ref{fig_intermeth}). An AHC with Ward's criterion was applied to this matrix in order to identify clusters of methods with similar cluster outputs. We visually identified five clusters of methods that had high ($> 0.7$) average within-cluster Rand index. The largest cluster included
classical AHCs such as \texttt{CompleteAbs}, \texttt{AverageAbs}, \texttt{WardAbs}, as well as the \texttt{Bic} and \texttt{BayesCov} methods proposed here. It should be noted that this class of algorithms generated solutions for this problem that were very close to one another (average
within-cluster Rand index $> 0.8$). The \texttt{BayesCorr} method was also close to that large group of methods, but not quite as much as the aforementioned methods (average Rand index of about 0.7), and was thus singled out as a separate cluster. The spectral methods were split into two clusters, depending on whether they were based on correlation (\texttt{SpecCorrAbs-1} and \texttt{SpecCorrAbs-10}) or partial correlation (\texttt{SpecCorrparAbs-1} and \texttt{SpecCorrparAbs-10}). Finally, the two variants of mutual information (\texttt{Infomut} and \texttt{InfomutNorm}) generated solutions that were highly similar to \texttt{SingleAbs}. It was reassuring that the variants of Bayes methods proposed here performed similarly to algorithms known to produce physiologically plausible solutions, such as Ward \gcite{Thirion-2014, Orban_P-sp}. While we found some analogy between \texttt{BayesCorr}, \texttt{BayesCov}, \texttt{Infomut} and \texttt{InfomutNorm}, it was intriguing that the variants of mutual information tested seemed to generate markedly different classes of solutions from the Bayes methods. We decided to examine the cluster solutions of these methods in more details.

As a reference, we examined the cluster solutions generated by \texttt{WardAbs}, in addition to two variants of Bayes clustering that yielded slightly different solutions (\texttt{BayesCov} and \texttt{BayesCorr}), and nomalized mutual information, \texttt{InfomutNorm}. To represent the typical behavior of each method across subjects, we generated a ``group'' consensus clustering summarizing the stable features of the ensemble of individual cluster solutions. This consensus clustering was generated by the evidence accumulation algorithm \gcite{Fred-2005} outlined below. First, each partition of each subject was represented as a binary 81-by-81 adjacency matrix $\matr{A} = ( A_{ i j } )$, where for each pair $( i, j )$ of brain regions, $A_{ i j } = 1$ if areas $i$ and $j$ were in the same cluster, and $A_{ i j } = 0$ otherwise. The adjacency matrices were then averaged across subjects and selected methods, yielding a 81-by-81 stability matrix $\matr{C} = ( C_{ i j } )$ where each element $C_{ij}$ coded for the frequency at which brain areas $i$ and $j$ fell in the same cluster. Finally this stability matrix was used as a similarity matrix in a AHC with Ward's criterion to generate one consensus partition. The brain regions were grouped into the same consensus cluster if they had a high probability of falling into the same cluster on average across subjects and methods, hence the
name consensus clustering.
\par
\begin{figure}[!htbp]
 \centering
 \includegraphics[width=\linewidth]{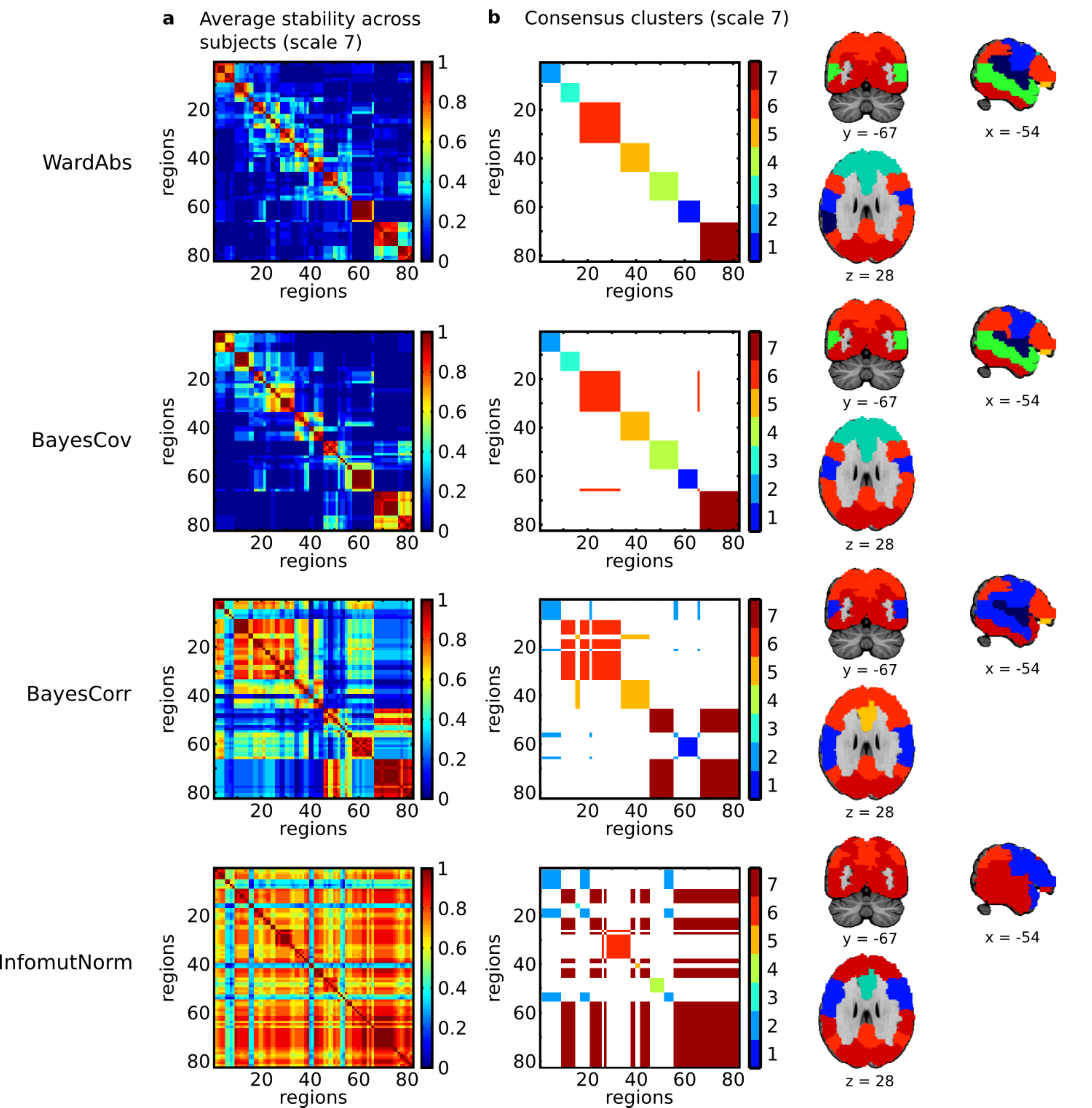}
 \caption{{\bf Real data---consensus clustering.} A consensus clustering was generated based on the average adjacency matrices across all subjects (Panel \textbf{a}). The (weighted) adjacency matrix associated with the consensus clustering is represented along with a volumetric brain parcellation (Panel \textbf{b}). The weights in the adjacency matrix were added to establish a visual correspondence with the volumetric representation. Note that the brain regions have been order based on the hierarchical clustering generated with \texttt{WardAbs}.}
\label{fig_consensus}
\end{figure}

Fig~\ref{fig_consensus} represents the stability matrices and consensus clusters, for the four methods of interest. As expected based on our first experiment on the similarity of cluster outputs, the \texttt{WardAbs} and \texttt{BayesCov} methods were associated with highly similar stability matrices and almost identical consensus clusters. Many areas of high consensus could be identified (with values close of 0 or 1), illustrating the very good agreement of the cluster solutions across subjects. The outline of the consensus clusters as well as a volumetric representation of the brain parcellation are presented in Fig~\ref{fig_consensus}b. Some of these clusters closely matched those reported in previous studies: cluster~7 can be recognized as being the visual network, cluster~2 the sensorimotor network, and clusters 6 and 3 the anterior and posterior parts of the default-mode network, respectively \gcite{Salvador-2005, van_den_Heuvel-2008, Bellec-2010}. By contrast with \texttt{WardAbs} and \texttt{BayesCorr}, the \texttt{InfomutNorm} tended to generate very large clusters, which was apparent both on the stability matrix and the consensus clusters. The \texttt{BayesCorr} method was intermediate between \texttt{BayesCov} and \texttt{InfomutNorm} in terms of cluster size. These decompositions into very large clusters do not fit current views on the organization of resting-state networks.

Overall, our analysis on real fMRI data led to the following conclusions. Three big subsets of methods emerged: spectral methods, mutual information (with \texttt{SingleAbs}), and finally all the other methods. Application of this last group of methods, which included the Bayes variants proposed here, resulted in a plausible decomposition into resting-state networks. In the absence of ground truth, it is not possible to further comment on the relevance of the differences in cluster solutions identified by the three groups of methods. We still noted that the methods based on mutual information led to large clusters that were difficult to interpret. Our interpretation is that the strategies implemented in \texttt{Infomut}  and \texttt{InfomutNorm} did not behave well for these datasets.
\par
As a final computational note, the time required by the different methods to cluster data is summarized in Table~\ref{tab:temps}. Although the differences in execution time may reflect the quality of the implementation, the methods proposed here were the slowest of the hierarchical methods, but were still faster than spectral clustering.

\begin{table}[!htbp]
 \centering
 \caption{{\bf Real resting-state fMRI data---computational cost.} Time required by each method to cluster one dataset. $^\star$ For nonhierarchical methods, we summed the times used to perform clustering at each scale.} \label{tab:temps}
 \begin{tabular}{llll}
  {\bf method} & {\bf minimum} & {\bf median} & {\bf maximum} \\
  \hline
  \texttt{CompleteAbs} & 10.9~ms & 11.5~ms & 24.5~ms \\
  \texttt{AverageAbs} & 11.0~ms & 11.6~ms & 25.2~ms \\
  \texttt{SingleAbs} & 11.1~ms& 11.7~ms & 49.5~ms \\
  \texttt{WardAbs} & 14.1~ms & 14.8~ms & 18.9~ms \\
  \texttt{InfomutNorm} & 159~ms & 170~ms & 261~ms \\
  \texttt{InfoMut} & 299~ms & 322~ms & 416~ms \\
  \texttt{Bic} & 666~ms & 673~ms & 810~ms \\
  \texttt{BayesCov} & 1.083~s & 1.094~s & 1.263~s \\
  \texttt{BayesCorr} & 1.108~s & 1.151~s & 1.133~s \\
  \texttt{SpecCorrparAbs-1}$^\star$ & 1.928~s & 2.065~s & 2.291~s \\
  \texttt{SpecCorrAbs-1}$^\star$ & 1.992~s & 2.176~s & 2.417~s \\
  \texttt{SpecCorrparAbs-10}$^\star$ & 13.251~s & 13.939~s & 14.301~s \\
  \texttt{SpecCorrAbs-10}$^\star$ & 14.456~s & 15.381~s & 16.101~s
 \end{tabular}
\end{table}

\section{Discussion}

\subsection{Contributions}

\paragraph{Summary.}

We here proposed some novel similarity measures well suited for the agglomerative hierarchical clustering of dependent variables. These measures rely on a Bayesian model comparison for multivariate normal random variables. On synthetic data with a known (ground truth) partition, hierarchical clustering based on the Bayesian measures was found to outperform several standard clustering procedures in terms of adjusted Rand index and classification accuracy. On the toy example, the Bayesian approaches led to result similar to those of mutual information clustering techniques, with the advantage of an automated thresholding. On the real fMRI data, the Bayesian measures led to results consistent with standard clustering methods, in contrast to methods based on mutual information, which exhibited a highly atypical behavior.

\paragraph{Bayesian normalization of mutual information.}

A key feature of the Bayesian approach is its ability to take the dimension of the clusters into account. Dimensionality is an important issue in two respects (see Appendix~\ref{ann:homog} for an illustration). First, mutual information is an extensive measure that depends on the dimension of the variables. This has motivated the introduction of a normalization factor in the application of mutual information to hierarchical clustering \gcite{Kraskov-2005, Kraskov-2009}. A second issue is the existence of a bias in the estimation of mutual information. This bias mechanically increases with the dimensionality of the variables. Because of the two issues described above, hierarchical clustering based on mutual information will tend to cluster unrelated but large variables rather than correlated variables of lower dimensions. As demonstrated on real fMRI data, the heuristic proposed by \gcite{Kraskov-2005} does not provide a general solution to the issue of dimensionality. Furthermore, it
removes some interesting features of mutual information, such as the additivity of the clustering measure. By contrast, the Bayesian approach takes the dimensionality into account in a principled way, providing a quantitative version of Occam's factor \gcite[Chap.~20]{Jaynes-2003}. The Bayesian normalization is additive rather than multiplicative, thus preserving the additive properties of mutual information.

\paragraph{From similarity measure to log Bayes factor.}

We defined the similarity measure $s ( \vect{X}_i, \vect{X}_j )$ between any two pairs of variables $\vect{X}_i$ and $\vect{X}_j$ as a log Bayes factor. At each step, the pair $( \vect{X}_{ i_0 }, \vect{X}_{ j_0 } )$ that had the largest similarity measure was merged. Taking into account the unique features of $s$ as the log Bayes factor defined in Equation~(\ref{eq:s}) allowed us to have access to a global measure of fit as defined in Equation~(\ref{eq:global:local}) as well as to provide an automatic stopping rule that behaved very well on simulated data. Going from a similarity measure to a log Bayes factor has other advantages that could take the clustering proposed here even further (see below).

\paragraph{Practical value of the \texttt{Bayes}/\texttt{Bic} clustering in fMRI.}

The new alternatives to mutual information introduced in this paper (i.e., \texttt{Bayes} and \texttt{Bic}) proved useful for the analysis of resting-state fMRI. The benefits were particularly clear when compared to \texttt{InfomutNorm}, which tended to create large, inhomogeneous clusters. By contrast, both \texttt{Bayes} and \texttt{Bic} had a behavior similar to standard hierarchical clustering based on Pearson's linear correlation. The possible benefits of \texttt{Bayes} and \texttt{Bic} over those canonical methods are still substantial. The mutual information first provides a multivariate measure of interaction that is well adapted to hierarchical brain decomposition \gcite{Tononi-1994, Marrelec-2008} and which has a clear interpretation in information theory \gcite{Watanabe-1960, Joe-1989b, Studeny-1998b}. For these reasons, the mutual information for Gaussian variables is more appealing than a simple average of pairwise correlation coefficients across clusters. Because mutual information is measured between clusters, it is natural to build the clusters themselves based on this metric. A second benefit of the proposed approach is that \texttt{Bic} proved to be the most stable of all tested methods in the range of 5--15 clusters on real fMRI datasets. The increase in stability over \texttt{Ward}'s was modest, but significant. This advantage may become even more substantial if the clustering is performed in higher dimension, i.e., with smaller areas than the AAL brain parcellation or even at the voxel level. 

\paragraph{Similarity vs. distance.}

Clustering techniques are based on either a similarity measure or a distance measure. While the description of the present manuscript mostly relied on the notion of similarity, going from one concept to the other one can generally be done with minor changes. For instance, standard hierarchical procedures which rely on the minimization of a distance to perform clustering (e.g., single, complete and average linkage) can be applied to cases where closeness is quantified by a measure of similarity, simply by using the opposite of the similarity matrix as a distance matrix. Although the resulting measure may not define a mathematically valid distance, it is not required for the procedure to work. Similarly, in a Bayesian framework, switching from a similarity measure to a distance measure tantamounts to switching from 
\begin{equation*}
 s ( \vect{X}_i,\vect{X}_j ) = \ln \frac{ \pr ( \matr{S}_{ i \cup j } | \mathcal{M}_D ) } { \pr ( \matr{S}_{ i \cup j } | \mathcal{M}_I ) }
\end{equation*}
to
\begin{equation*}
 d ( \vect{X}_i,\vect{X}_j ) = - s ( \vect{X}_i,\vect{X}_j ) = \ln \frac{ \pr ( \matr{S}_{ i \cup j } | \mathcal{M}_I ) } { \pr ( \matr{S}_{ i \cup j } | \mathcal{M}_D ) },
\end{equation*}
that is, from the log ratio of the marginal model likelihoods in favor of dependent variables to the log ratio of the marginal model likelihoods in favor of \emph{in}dependent variables.

\subsection{Modeling choices}

\paragraph{Choice of priors.}

Any Bayesian analysis requires the introduction of prior distributions. In the present study, we needed the prior distribution for the covariance matrix associated to any clustering of $\vect{X}$. Our choices were guided by one assumption, one rule of consistency, and one rule of simplicity. First, our
assumption was to not assume a priori any sort of dependence between covariance matrices associated to different clusters. This allowed for the decomposition of any prior as a product of independent priors, see Equations (\ref{eq:indep:prior:1}) and (\ref{eq:part:prior:1}). The rule of consistency was to consider that the prior for a given covariance matrix should not be contradictory at different levels of the hierarchy. This is why we set the prior distribution for the global covariance matrix $\matr{\Sigma}$ as an inverse-Wishart distribution and then derived the prior for any covariance matrix $\matr{\Sigma}_k$ as the prior distribution for $\matr{\Sigma}$ integrated over all parameters that do not appear in $\matr{\Sigma}_k$; using such an approach, the resulting prior turned out to be an inverse-Wishart distribution as well. Last, the rule of simplicity is the one that dictated the use of inverse-Wishart distributions as prior distributions for the covariance matrices. Such a family of priors had the twofold advantage of being closed by marginalization and allowing for closed-form expressions of the quantities of interest. An inverse-Wishart distribution is characterized by two parameters: the degrees of freedom $\nu$ and the inverse scale matrix $\matr{\Lambda}$. If $\matr{\Sigma}$ is a $D$-by-$D$ matrix, we must have $\nu > D - 1$ for the distribution to be normalized. Also, $\matr{\Lambda}$ must be positive definite. From there, we had two strategies. The first one was to set the degree of freedom to the lowest value that still corresponded to a well defined distribution ($\nu = D$) and a diagonal scale matrix that optimized the marginal model likelihood. An alternative approach was to work with the sample correlation matrix, set $\nu = D + 1$ and equal $\matr{\Lambda}$ with the $D$-by-$D$ identity matrix $\matr{I}$, since this choice corresponds to a distribution that is associated with uniform marginal distributions of the
correlation coefficients \gcite{Barnard_J-2000}. While we believe that our assumption and the rule of consistency are sensible choices, we must admit that we are not quite as content with the choice of inverse-Wishart distributions for priors. The major issue with such a family of priors is that they simultaneously constrain the structure of correlation and the variances. By contrast, it seems
intuitive that clustering should depend on the correlation structure only, not on the variances. As such, the prior on the variances should ideally be set independently from the correlation structure. Priors that separate variance and correlation have already been proposed \gcite{Barnard_J-2000}. Unfortunately, the use of such priors would make it impossible to provide a closed form for the marginal model likelihood. While numeric schemes could be implemented to circumvent this issue, it would render the procedure proposed much more complex and computationally burdensome. By contrast, the algorithm detailed here is rather straightforward. Also, the influence of the prior vanishes when the sample size increases. From a practical perspective, the three methods proposed here (two, \texttt{BayesCov} and \texttt{BayesCorr}, with different priors and one, \texttt{Bic}, not influenced by the prior) exhibited similar behaviors and still outperformed other existing methods in the
simulation study. We take it as a proof of the robustness of the method to the choice of prior.

\paragraph{Covariance vs. precision matrix modeling.}

The presence of clusters of variables that are mutually independent is equivalent to having a covariance matrix $\matr{\Sigma}$ that is block diagonal, which is itself equivalent to having a precision (or concentration) matrix $\matr{\Upsilon} = \matr{\Sigma} ^ { - 1 }$ that is also block diagonal. One could therefore solve the problem using precision matrices instead of covariance matrices. The corresponding calculations can be found in Appendix~\ref{ann:conc}. The main difference between the two approaches stems from the fact that a submatrix of the inverse covariance matrix is not equal to the inverse of the corresponding submatrix of the covariance matrix, that is, $( \matr{\Lambda} ^ { - 1 } )_k \neq ( \matr{\Lambda}_k ) ^ { - 1 }$, unless $\matr{\Lambda}$ is block diagonal; also, Wishart and inverse-Wishart distributions do not marginalize the same way. These differences are to be related to the fact that a submatrix of a covariance matrix is better estimated than the whole covariance matrix, while the same does not hold for a precision matrix. From there, we could expect the covariance-based approach to perform better than the precision-based approach, and the difference to increase with increasing $D$. This was confirmed on our synthetic data, where the precision-based approach behaved as well as \texttt{BayesCov} and \texttt{BayesCorr} for $D = 6$ but had worse results than these two approaches for $D \in \{ 10, 20, 40 \}$. Besides, performance of the automated stopping rule was much more efficient with \texttt{BayesCov} and \texttt{BayesCorr} than it was with the precision-based approach. As a final note, basing the method on concentration matrices yielded a slower algorithm, arguably because of the matrix inversions that are required.

\paragraph{Sample covariance matrix vs. full dataset.}

Intuitively, the structure of dependence of a multivarite normal distribution is included in its covariance matrix. All existing algorithms that we used here do not need the full dataset but only the sample covariance (or correlation) matrix. This is why we started with a likelihood model that only considers the covariance matrix [see Equations (\ref{eq:dep:vrais}) and (\ref{eq:indep:vrais})]. Rigorously, this model is only valid for $N \geq D$; for $N < D$, one should resort to a model of the full data as being multivariate normal with a mean $\vect{\mu}$ and covariance matrix $\matr{\Sigma}$. Nonetheless, we kept the 'intuitive' approach as it has fewer hyperparameters, is easier to deal with, and leads to formulas that are simpler to interpret. Also, the resulting similarity measure [Equation~(\ref{eq:bayes:def:2a})] is not restricted to $N \geq D$, but is well defined for $N < D$ as well. From a practical perspective, the 'intuitive' algorithm only requires the sample covariance matrix as input, while a full model would also require the sample mean. Finally, this simpler model exhibited good behavior on our synthetic data, even for small datasets.

\subsection{Directions for future work}

\paragraph{Computational costs.}

Regarding the computational cost, measures derived from mutual information or a Bayesian approach are more demanding than standard methods such as \texttt{Average} or \texttt{Ward}. The derivation of the similarity matrix and the search for the most similar pairs of clusters are the two critical operations that can be optimized to decrease computation time. It is always possible to speed up these two steps by taking advantage of the fact that the similarity matrices of two successive iterations of the algorithm have many elements in common, as all but one element of a partition at a given iteration are identical to the elements of the partition of the previous iteration. Critically, in the case of \texttt{Average} and \texttt{Ward} methods, it is in addition possible to derive the similarity matrix at every iteration only based on the similarity matrix at initialization through successive updates using the Lance-Jambu-Williams recursive formula \gcite{Batagelj-1988}. By contrast, other measures, including \texttt{BayesCov}, \texttt{BayesCorr}, \texttt{Bic}, \texttt{Infomut}, and \texttt{InfomutNorm}, have to be re-evaluated independently at every step, which means that the determinant of a covariance matrix with increasing size has to be computed. Finding an update formula analogous to Lance-Jambu-Williams for clustering methods based on variants of the mutual information would substantially accelerate these algorithms.

\paragraph{From deterministic to stochastic clustering.}

Another extension would be to replace the deterministic rule of selecting the pair with largest similarity measure for merging by a probabilistic rule where the probability to cluster a given pair is given by the posterior probability of the resulting clustering.

\paragraph{Group analysis.}

A last extension that could be easily implemented in the present framework is the generalization of the method to account for $E$ different entities (e.g., subjects in fMRI) sharing the same structure but with potentially different covariance matrices for that structure. If each entity $e$ is characterized by a variable $\vect{X} ^ { [ e ] }$ and corresponding sample sum-of-square matrix $\matr{S} ^ { [ e ] }$, one can perform AHC on each $\vect{X} ^ { [ e ] }$ using $\matr{S} ^ { [ e ] }$ and the corresponding similarity measure $s ^ { [ e ] }$. However, with a straightforward modification of the present method, one could also perform global AHC of all $E$ covariance matrices considered simultaneously. Assuming that the covariance matrices of the different elements are independent given the common underlying structure, then the resulting similarity measure is the sum of all individual similarity measures $s ^ { [ e ] }$'s.

\paragraph{Generalization to other types of distribution.}

Altogether, the Bayesian framework that we used provides a principled way to generalize our approach to distributions other than multivariate normal ones. Such generalization would potentially account for a wide variety of situations, such as nonlinear dependencies or discrete distributions. This would widen the scope of possible applications of the technique, e.g., genetics \gcite{Butte-2000, Zhou_Xiaobo-2004, Dawy-2006, Priness-2007}. The issues related to this generalization are twofold. First, one needs a model of dependence. In the discrete case, one could think of multinomial distributions, originating from categorical, i.e., generalized Bernoulli distributions \gcite[\S3.4]{Papoulis-1965}. In the continuous case, multivariate normal distributions are a first choice model beyond which it is not clear what to use. Multivariate Student-$t$ distributions could be considered, even though our results on simulated data would tend to hint that the difference with multivariate normal distriubtions might not be that large. One could also consider using models where dependence is controled independently of the marginal distributions, such as multivariate copulas \gcite{Nelsen-1999, Fischer-2011}. Another issue is the possibility to express the marginal posterior likelihood of the data given the model selected. For multivariate discrete variables, we expect it to be feasible, albeit computationally very challenging and sensitive to the type of prior distribution. For other distributions, obtaining a closed form might be out of reach. Nonetheless, the marginal posterior likelihood could be approximated using various criteria, such as the AIC \gcite{Akaike-1974} or variants thereof---AICc \gcite{Burnham-2004}; \gcite[\S2.3.1]{McQuarrie-1998} or AICu \gcite[\S2.4.1]{McQuarrie-1998}---, or the BIC \gcite{Schwarz-1978}, which naturally appeared in the present derivation. In any case, any approach beyond multivariate normal distributions would drastically increase the complexity of our approach, both in terms of inference and computation.

\paragraph{Application to truly hierarchical data.}

In the present manuscript, we used a hierarchical algorithm as a way to extract an underlying structure of dependence from data. Our assumption was that there was one such structure. Such an approach provided a simple and efficient clustering algorithm with an interesting connection to mutual information. However, the method as presented here is not able to deal with data that are truly organized hierarchically. Extending it to deal with such data would improve the scope of the algorithm. One way to do would be to use Dirichlet process mixtures \gcite{Heller_KA-2005, Heller_KA-2007, Savage_RS-2009, Cooke_EJ-2011, Darkins-2013, Sirinukunwattana-2013}, together with a model of dependent variables.

\section{Conclusion}

In this paper, we proposed a procedure based on Bayesian model comparison to decide whether or not to merge Gaussian multivariate variables in an agglomerative hierarchical clustering procedure. The resulting similarity measure was found to be closely related to the standard mutual information, with some additional corrections for the dimensionality of the datasets. These new Bayesian alternatives to mutual information turned out to be beneficial to hierarchical clustering on Gaussian simulations and real datasets alike. Because of the simplicity of its implementation, its good practical performance and the potential generalizations to other types of random variables, we believe that the approach presented here is a useful new tool in the context of hierarchical clustering.


\section*{Acknowledgments}

The authors are grateful to the members of the 1000 functional connectome consortium for publicly releasing the 'Atlanta' data sample, to the UNF (Unit{\'e} de neuroimagerie fonctionnelle, Universit{\'e} de Montr{\'e}al, Montr{\'e}al, Qc, Canada) for providing them with computational resources, to Mathieu Desrosiers (UNF) for his expertise and suppport running the simulations on the grid engine. Part of this study was performed while G.M. was at the UNF.

\section*{Author contributions}

G.M. and P.B. conceived and designed the experiments, performed the experiments, and analyzed the data. G.M., A.M., and P.B. contributed reagents/materials/analysis tools and wrote the manuscript. G.M. conceived the theoretical model.

\section*{Competing Interests}

P.B. offers consulting in brain image analysis for two Contract Research Organizations, NeuroRX and Biospective, Montreal, Canada.

\section*{Funding}

A.M. is supported by Deutsche Forschungsgemeinschaft (DFG) grant SFB 936/Z3. P.B. is supported by the Fonds de Recherche du Qu{\'e}bec -- Sant{\'e} (FRQS). The funders had no role in study design, data collection and analysis, decision to publish, or preparation of the manuscript

\appendix

\section{Illustration of the behavior of estimated mutual information} \label{ann:homog}

In the case of a $D$-dimensional variable $\vect{X}$ with a multivariate normal distribution, the mutual information between two subvectors $\vect{X}_i$ and $\vect{X}_j$ is given by
\begin{equation} \label{eq:im:normal}
 I ( \vect{X}_i, \vect{X}_j ) = \frac{ 1 } { 2 } \ln \frac{ | \matr{\Sigma}_i | \, | \matr{\Sigma}_j | } { | \matr{\Sigma}_{ i \cup j } | },
\end{equation}
where $\matr{\Sigma}_k$, $k \in \{ i, j, i \cup j \}$, is the covariance matrix of $\vect{X}_k$  and $| \cdot |$ is the usual determinant function. 

\paragraph{Estimation bias.}

In this case, mutual information is estimated using its plug-in estimator $\hat{I}$, that is, Equation~(\ref{eq:im:normal}) where the model covariance matrices have been replaced by their estimators (i.e., the corresponding sample covariance matrices). For large $N$, this estimator suffers from a systematic bias, as it was shown that \gcite{Marrelec-2011}
\begin{equation*}
 \esp ( \hat{I} ) = I ( \vect{X}_i, \vect{X}_j ) + \frac{ D_i D_j } { 2 N } + O \left( \frac{ 1 } { N ^ 2 } \right). 
\end{equation*}
This bias is additive and is a (quadratic) function of the problem dimensionality. This is consistent with the fact that $2 N \, I ( \vect{X}_i, \vect{X}_j )$ asymptotically follows a chi square distribution with $D_i D_j$ degrees of freedom \gcite[Chap.~12, \S3.6]{Kullback-1968};
\gcite[\S11.3.2]{Press_SJ-2005}.

\paragraph{Extensivity of the measure.}

Besides the above problem of estimation, mutual information itself is an extensive quantity i.e., it mechanically increases with an increase in the dimensionality of the problem. To better demonstrate this effect, consider a model of homogeneous matrix for $\matr{\Sigma}$. More specifically, let
$\matr{A}_D ( \rho )$ be a $D$-by-$D$ homogeneous matrix with parameter $\rho$, i.e., a matrix with 1s on the diagonal and all off-diagonal elements equal to $\rho$. $\matr{A}_D ( \rho )$ has two eigenvalues: $1 + ( D - 1 ) \rho$ with multiplicity 1 (associated with the vector composed only of 1s) and $1 - \rho$ with multiplicity $D - 1$ (associated with the subspace of vectors with a zero mean). This covariance matrix is therefore positive definite for $0 \leq \rho < 1$ and its determinant is given by $[ 1 + ( D - 1 ) \rho ] ( 1 - \rho ) ^ { D - 1 }$. Assuming that $\matr{\Sigma} = \matr{A}_D ( \rho )$, mutual information can be expressed as
\begin{equation}
 I ( \vect{X}_i, \vect{X}_j ) = \frac{ 1 } { 2 } \ln \frac{ [ 1 + ( D_i - 1 ) \rho ] [ 1 + ( D_j - 1 ) \rho ] } { ( 1 - \rho ) [ 1 + ( D_i + D_j - 1 ) \rho ] },
\end{equation}
where $D_k$, $k \in \{ i, j, i \cup j \}$, is the dimension of $\vect{X}_k$. This expression can lead to counter-intuitive results: the mutual information is about 0.31 for $D_i = D_j = 5$ and $\rho = 0.3$, while it is about 0.33 for $D_i = D_j = 7$ and $\rho = 0.25$. More generally, $I ( \vect{X}_i, \vect{X}_j )$ is an increasing function of $D_i$ and $D_j$ (as can be seen by differentiation). This means that, for the same value of $\rho$, mutual information will be larger for larger values of $D_i$ and $D_j$. It also means that mutual information will favor the merging of two variables with smaller marginal correlations if the dimensionality of the variables is large enough, thus systematically favoring larger clusters. To give a feeling of the amplitude of this behavior, set $D_k' = D_k - 1$ and assume that we have $D_k' \rho \gg 1$. Mutual information can then be approximated by
\begin{equation*}
 I ( \vect{X}_i, \vect{X}_j ) \approx \frac{ 1 } { 2 } \ln \frac{ D_i' D_j' \rho } { ( 1 - \rho ) D_{ i \cup j }' } = \frac{ 1 } { 2 } \ln \frac{ \rho } { 1 - \rho } + \frac{  1 } { 2 } \ln \frac{ D_i' D_j' } { D_{ i \cup j }' }.
\end{equation*}

\section{Asymptotic form of the log Bayes factor} \label{ann:asympt}

Assume that $N \to \infty$. Starting from the expression $\phi$ of Equation~(15), we have
\begin{equation} \label{eq:phi:2}
 \phi ( N + \nu_k, \matr{\Lambda}_k + \matr{S}_k ) = - \frac{ N + \nu_k } { 2 } \ln | \matr{\Lambda}_k + \matr{S}_k | + \sum_{ d = 1 } ^ { D_k } \ln \Gamma \left( \frac{ N + \nu_k + 1 - d } { 2 } \right).
\end{equation}
Defining $\wh{\matr{S}}_k$ as the standard sample covariance matrix, i.e., $\matr{S}_k = N \wh{\matr{S}}_k$, the first term of the right-hand side can be expanded as
\begin{eqnarray*}
 \frac{ N + \nu_k } { 2 } \ln | \matr{\Lambda}_k + \matr{S}_k | & = & \frac{ N + \nu_k } { 2 } \ln | \matr{\Lambda}_k + N \wh{\matr{S}}_k | \\
 & = &  \left( \frac{ N } { 2 } + \frac{ \nu_k } { 2 } \right) \left[ D_k \ln N + \ln | \wh{\matr{S}}_k | + \ln | \matr{I} + ( N \wh{\matr{S}}_k ) ^ { - 1 } \matr{\Lambda}_k | \right] \\
 & = & \frac{ D_k N } { 2 } \ln N + \frac{ N } { 2 } \ln | \wh{\matr{S}}_k | + \frac{ D_k \nu_k } { 2 } \ln N + O ( 1 ),
\end{eqnarray*}
since $| a \matr{A} | = a ^ { \dim ( \matr{A} ) } | \matr{A} |$ for any positive number $a$ and matrix $\matr{A}$. In the expression of $\phi$ in Equation~(\ref{eq:phi:2}), each term in the sum can be approximated using Stirling approximation \gcite[p.~257]{Abramowitz-1972}
\begin{equation*}
 \ln \Gamma ( z ) = \left( z - \frac{ 1 } { 2 } \right) \ln z - z + O ( 1 ). 
\end{equation*}
Setting $z = ( N + \nu_k + 1 - d ) / 2$ for $l \in \{ i, j, i \cup j \}$, we obtain
\begin{equation*}
 \ln \Gamma \left( \frac{ N + \nu_k + 1 - d } { 2 } \right) = \frac{ N + \nu_k - d } { 2 } \ln N - \frac{ N } { 2 } ( 1 + \ln 2 ) + O ( 1 ).
\end{equation*}
Summing this expression over $d = 1, \dots, D_k$ and using the fact that $\sum_{ d = 1 } ^ { D_k } d = D_k ( D_k + 1 ) / 2$ leads us to
\begin{eqnarray*}
 \sum_{ d = 1 } ^ { D_k } \ln \Gamma \left( \frac{ N + \nu_k + 1 - d } { 2 } \right) & = & D_k \left[ \frac{ N + \nu_k } { 2 } \ln N - \frac{ N } { 2 } ( 1 + \ln 2 ) \right] \\
 & & \qquad - \frac{ D_k ( D_k + 1 ) } { 4 } \ln N + O ( 1 ). \\
\end{eqnarray*}
We then have for $\phi$
\begin{equation} \label{eq:phi:dl}
 \phi ( N + \nu_k, \matr{\Lambda}_k + \matr{S}_k ) = - \frac{ N } { 2 } \ln | 
\wh{\matr{S}}_k | - \frac{ D_k N } { 2 } ( 1 + \ln 2 ) - \frac{ D_k ( D_k + 1 )
} { 4 } \ln N + O ( 1 ).
\end{equation}
Since $\phi ( \nu_k, \matr{\Lambda}_k )$ does not depend on $N$, it is $O ( 1 )$, and the Taylor expansion of $\Delta \phi_k$, as defined in Equation~(14), is also given by Equation~(\ref{eq:phi:dl}). Finally, the approximation for $s ( \vect{X}_i, \vect{X}_j )$ of Equation~(11) is given by
\begin{eqnarray} \label{eq:bic}
 s ( \vect{X}_i, \vect{X}_j ) & = & \frac{ N } { 2 } \ln \frac{ | \wh{\matr{S}}_i | \, | \wh{\matr{S}}_j | } { | \wh{\matr{S}}_{ i \cup j } | } - \frac{ 1 } { 2 } \left[ \frac{ D_{ i \cup j } ( D_{ i \cup j } + 1 ) } { 2 } - \sum_{ k \in \{ i, j \} } \frac{ D_k ( D_k + 1 ) } { 2 } \right] \ln N + O ( 1 ) \nonumber \\
 & = & N \, \hat{I} ( \vect{X}_i, \vect{X}_j ) - \frac{ D_i D_j } { 2 } \ln N + O ( 1 ), 
\end{eqnarray}
where we used the fact that $D_{ i \cup j } = D_i + D_j$, and where we set
\begin{equation*}
 \hat{I} ( \vect{X}_i, \vect{X}_j ) = \frac{ 1 } { 2 } \ln \frac{ | \wh{\matr{S}}_i | \, | \wh{\matr{S}}_j | } { | \wh{\matr{S}}_{ i \cup j } | }.
\end{equation*}

\section{Hyperparameter optimization} \label{ann:optim}

Given a clustering $\{ \vect{X}_1, \dots, \vect{X}_K \}$ of $\vect{X}$, optimizing  the marginal likelihood with respect to a diagonal $\vect{\Lambda}$ leads to the optimization of
\begin{equation*}
 \sum_{ d = 1 } ^ D \frac{ \nu - D + D_{ k_d } } { 2 } \ln \Lambda_{ d d } - \sum_{ k = 1 } ^ K \frac{ N + \nu - D + D_k } { 2 } \ln | \matr{S}_k + \matr{\Lambda}_k |,
\end{equation*}
where $k_d$ is the cluster containing $X_d$. Differentiation with respect to $\Lambda_{ d d }$ leads to
\begin{equation} \label{eq:ann:opt}
 \frac{ \nu - D + D_{ k_d } } { 2 \Lambda_{ d d } } - \frac{ N + \nu - D + D_{ k_d } } { 2 } \left[ ( \matr{S}_{ k_d } + \matr{\Lambda}_{ k_d } ) ^ { - 1 } \right]_{ d d } = 0.
\end{equation}
To obtain the solution of this equation, we notice that the equation is equivalent to
\begin{equation*}
 \Lambda_{ d d } = \frac{ \nu - D + D_{ k_d } } { N + \nu - D + D_{ k_d } } \left\{ \left[ ( \matr{S}_{ k_d } + \matr{\Lambda}_{ k_d } ) ^ { - 1 } \right]_{ d d } \right\} ^ { - 1 }.
\end{equation*}
Optimizing at the first level (i.e., with $D$ clusters and $D_{ k_d } = 1$) yields
\begin{equation*}
 \Lambda_{ d d } = \frac{ \nu - D + 1 } { N } S_{ d d }.
\end{equation*}

\section{Real fMRI data: datasets and preprocessing} \label{ann:irmf}

We used the 'Atlanta' resting-state fMRI database \gcite{Liu_Hesheng-2009}. This resource was made publicly available as part of the 1000-connectome project\footnote{\texttt{http://www.nitrc.org/projects/fcon\_1000/}} \gcite{Biswal-2010}. The Atlanta sample includes 28 subjects (age ranging from 22 to 57 years, 15 women) with one structural MRI and one fMRI run each (205 volumes, TR $=$ 2~s), acquired on a 3~T scanner. The datasets were preprocessed using the neuroimaging analysis kit\footnote{\texttt{http://wiki.bic.mni.mcgill.ca/index.php/NiakFmriPreprocessing}} (NIAK), version 0.6.5c \gcite{Bellec-2012}. The parameters of a rigid body motion were first estimated at each time frame of the fMRI dataset (no correction of inter-slice difference in acquisition time was applied). The median volume of the fMRI time series was coregistered with a T\textsubscript{1} individual scan using
Minctracc\footnote{\texttt{http://wiki.bic.mni.mcgill.ca/index.php/MinctraccManPage}} \gcite{Collins_DL-1994}, which was itself transformed to the Montreal Neurological Institute (MNI) non-linear template \gcite{Fonov-2011} using the CIVET\footnote{\texttt{http://wiki.bic.mni.mcgill.ca/index.php/CIVET}} pipeline
\gcite{Zijdenbos-2002}. The rigid-body transform, fMRI-to-T\textsubscript{1} transform and T\textsubscript{1}-to-stereotaxic transform were all combined, and the functional volumes were resampled in the MNI space at a 3~mm isotropic resolution. The ``scrubbing'' method of \gcite{Power-2012} was used to remove the volumes with excessive motion (frame displacement greater than 0.5). The following nuisance parameters were regressed out from the time series at each voxel: slow time drifts (basis of discrete cosines with a 0.01~Hz high-pass cut-off), average signals in conservative masks of the white matter and the lateral ventricles as well as the first principal components (95\% energy) of the six rigid-body motion parameters and their squares \gcite{Giove-2009}. The fMRI volumes were then spatially smoothed with a 6~mm isotropic Gaussian blurring kernel. Because some of the measures considered in this paper are poorly conditioned when the number of spatial locations is larger than the number of time points
(\texttt{BIC}, \texttt{Infomut} and \texttt{InfomutNorm}), the fMRI time series were spatially averaged on each of the areas of the AAL brain template \gcite{Tzourio-Mazoyer-2002}. To further reduce the spatial dimension, only the 81 cortical AAL areas were included in the analysis (excluding the cerebellum, the basal ganglia and the thalamus). The clustering methods were applied to these regional time series. Note that 8 subjects were excluded because there was not enough time points left after scrubbing (a minimum number of 190 volumes was selected as acceptable), and one additional subject had to be excluded because the quality of the T\textsubscript{1}-fMRI coregistration was substandard (by visual inspection). A total of 19 subjects was thus actually used in this analysis.

\section{Model comparison with the concentration matrix} \label{ann:conc}

\subsection{Hypothesis of dependence}

$\matr{S}_{i \cup j}$ is Wishart distributed with $N$ degrees of freedom and scale matrix $\matr{\Sigma}_{i \cup j} = \matr{\Upsilon}_{i \cup j} ^ { - 1 }$ 
\begin{equation*}
 \pr ( \matr{S}_{i \cup j} | {\mathcal M}_D, \matr{\Upsilon}_{i \cup j} ) = \frac{ | \matr{S}_{ i \cup j } | ^ { \frac{ N - D_{i \cup j} - 1 } { 2 } } } { Z ( D_{i \cup j}, N ) } \, | \matr{\Upsilon}_{i \cup j} | ^ { \frac{ N } { 2 } } \exp \left[ - \frac{ 1 } { 2 } \tr ( \matr{\Upsilon}_{i \cup j} \matr{S}_{i \cup j} ) \right].
\end{equation*}
The prior for $\matr{\Sigma}_{ i \cup j }$ [Equation~(4) of the manuscript] directly translates into a prior for $\matr{\Upsilon}_{ i \cup j }$ that is Wishart with $\nu_{ i \cup j }$ degrees of freedom and scale matrix $\matr{\Omega}_{ i \cup j } = \matr{\Lambda}_{ i \cup j } ^ { - 1 }$
\begin{equation*}
 \pr ( \matr{\Upsilon}_{i \cup j} | {\mathcal M}_D ) = \frac{ | \matr{\Omega}_{ i \cup j } | ^ { - \frac{ \nu_{ i \cup j } } { 2 } } } { Z ( D_{i \cup j}, \nu_{ i \cup j } ) } | \matr{\Upsilon}_{i \cup j} | ^ { \frac{ \nu_{ i \cup j } - D_{ i \cup j } - 1 } { 2 } } \exp \left[ - \frac{ 1 } { 2 } \tr ( \matr{\Omega}_{ i \cup j } ^ { - 1 } \matr{\Upsilon}_{i \cup j} ) \right].
\end{equation*}
This leads to a marginal likelihood of
\begin{eqnarray*}
 \pr ( \matr{S}_{i \cup j} | {\mathcal M}_D ) & = & \frac{ | \matr{S}_{ i \cup j } | ^ { \frac{ N - D_{i \cup j} - 1 } { 2 } } | \matr{\Omega}_{ i \cup j } | ^ { - \frac{ \nu_{ i \cup j } } { 2 } } } { Z ( D_{i \cup j}, N ) \, Z ( D_{i \cup j}, \nu_{ i \cup j } ) } \\
 & & \qquad \times \int | \matr{\Upsilon}_{i \cup j} | ^ { \frac{ N + \nu_{ i \cup j } - D_{ i \cup j } - 1 } { 2 } } \exp \left\{ - \frac{ 1 } { 2 } \tr \left[ \matr{\Upsilon}_{i \cup j} (  \matr{S}_{i \cup j} + \matr{\Omega}_{i \cup j} ^ { - 1 } ) \right] \right\} \, \ud \matr{\Upsilon}_{ i \cup j }.
\end{eqnarray*}
The integrand is proportional to a Wishart distribution with $N + \nu_{ i \cup j }$ degrees of freedom and scale matrix $(  \matr{S}_{i \cup j} + \matr{\Omega}_{i \cup j} ^ { - 1 } ) ^ { - 1 }$, leading to
\begin{equation} \label{eq:ann:dep}
 \pr ( \matr{S}_{i \cup j} | {\mathcal M}_D ) = \frac{ | \matr{S}_{ i \cup j } | ^ { \frac{ N - D_{i \cup j} - 1 } { 2 } } } { Z ( D_{i \cup j}, N ) } \frac{ Z ( D_{ i \cup j }, N + \nu_{ i \cup j} ) } { Z ( D_{i \cup j}, \nu_{ i \cup j } ) } \frac { \left| \left( \matr{S}_{i \cup j} + \matr{\Omega}_{ i \cup j } ^ { - 1 } \right) ^ { - 1 } \right| ^ { \frac{ N + \nu_{ i \cup j } } { 2 } } } { | \matr{\Omega}_{ i \cup j } | ^ { \frac{ \nu_{ i \cup j } } { 2 } } }.
\end{equation}

\subsection{Hypothesis of independence}

In the case of independence, the same likelihood holds with the addition that, since $\matr{\Upsilon}_{i \cup j}$ is block diagonal with blocks $\matr{\Upsilon}_i$ and $\matr{\Upsilon}_j$, we have $| \matr{\Upsilon}_{i \cup j} |  = | \matr{\Upsilon}_i | \, | \matr{\Upsilon}_j |$ as well as $\tr ( \matr{\Upsilon}_{i \cup j} \matr{S}_{i \cup j} ) = \tr ( \matr{\Upsilon}_i \matr{S}_i ) + \tr ( \matr{\Upsilon}_j \matr{S}_i )$, leading to a likelihood of
\begin{equation*}
 \pr ( \matr{S}_{i \cup j} | {\mathcal M}_I, \matr{\Upsilon}_{i \cup j} ) = \frac{ | \matr{S}_{ i \cup j } | ^ { \frac{ N - D_{i \cup j} - 1 } { 2 } } } { Z ( D_{i \cup j}, N ) } \prod_{ k \in \{ i, j \} } | \matr{\Upsilon}_k | ^ { \frac{ N } { 2 } } \exp \left[ - \frac{ 1 } { 2 } \tr ( \matr{\Upsilon}_k
\matr{S}_k ) \right].
\end{equation*}
The prior for $\matr{\Upsilon}_k$ that derives from that of $\matr{\Upsilon}_{ i \cup j }$ is a Wishart distribution with $\nu_{i \cup j}$ degrees of freedom and scale matrix $\matr{\Omega}_k$ \gcite[\S5.1.2]{Press_SJ-2005}
\begin{equation*}
 \pr ( \matr{\Upsilon}_k | {\mathcal M}_I ) = \frac{ | \matr{\Omega}_k | ^ { - \frac{ \nu_{ i \cup j } } { 2 } } } { Z ( D_k , \nu_{ i \cup j } ) } | \matr{\Upsilon}_k | ^ { \frac{ \nu_{ i \cup j } - D_k - 1 } { 2 } } \exp \left[ - \frac{ 1 } { 2 } \tr ( \matr{\Omega}_k ^ { - 1 } \matr{\Upsilon}_k ) \right].
\end{equation*}
This leads to a marginal likelihood of
\begin{eqnarray*}
 \pr ( \matr{S}_{i \cup j} | {\mathcal M}_I ) & = & \frac{ | \matr{S}_{ i \cup j } | ^ { \frac{ N - D_{i \cup j} - 1 } { 2 } } | \matr{\Omega}_k | ^ { - \frac{ \nu_{ i \cup j } } { 2 } } } { Z ( D_{i \cup j}, N ) \, Z ( D_i, \nu_{ i \cup j } ) \, Z ( D_j, \nu_{ i \cup j } ) } \\
 & & \qquad \times \prod_{ k \in \{ i, j \} } \int | \matr{\Upsilon}_k | ^ { \frac{ N + \nu_{ i \cup j } - D_k - 1 } { 2 } } \exp \left\{ - \frac{ 1 } { 2 } \tr \left[ \matr{\Upsilon}_k ( \matr{S}_k + \matr{\Omega}_k ^ { - 1 } ) \right] \right\} \, \ud \matr{\Upsilon}_k.
\end{eqnarray*}
Each integrand is proportional to a Wishart distribution with $N + \nu_{ i \cup j }$ degrees of freedom and scale matrix $( \matr{S}_k + \matr{\Omega}_k ^ { - 1 } ) ^ { - 1 }$, leading to
\begin{equation} \label{eq:ann:indep}
 \pr ( \matr{S}_{i \cup j} | {\mathcal M}_I ) = \frac{ | \matr{S}_{ i \cup j } | ^ { \frac{ N - D_{i \cup j} - 1 } { 2 } } } { Z ( D_{i \cup j}, N ) } \prod_{ k \in \{ i, j \} } \frac{ Z ( D_k, N + \nu_{ i \cup j } ) } { Z ( D_k, \nu_{ i \cup j } ) } \frac{ \left| \left( \matr{S}_k + \matr{\Omega}_k ^ { - 1 } \right) ^ { - 1 } \right| ^ { \frac{ N + \nu_{ i \cup j } } { 2 } } } { | \matr{\Omega}_k | ^ {\frac{ \nu_{ i \cup j } } { 2 } } }.
\end{equation}


\end{document}